\documentclass{ieeeaccess}
\usepackage{cite}
\usepackage{amsmath,amssymb,amsfonts}
\usepackage{algorithm}
\usepackage{algpseudocode}
\usepackage{graphicx}
\usepackage{textcomp}
\usepackage{booktabs}
\usepackage{multirow}
\usepackage{array}
\usepackage{hyperref}
\usepackage{caption}
\usepackage{adjustbox}
\usepackage{comment}

\usepackage{soul}
\usepackage{xcolor}

\def\BibTeX{{\rm B\kern-.05em{\sc i\kern-.025em b}\kern-.08em
    T\kern-.1667em\lower.7ex\hbox{E}\kern-.125emX}}
\begin{document}
\history{Date of publication xxxx 00, 0000, date of current version xxxx 00, 0000.}
\doi{}

\title{Customer Churn Prediction on Structured Data Using FT-Transformer and Stacking Ensembles}

\author{\uppercase{Joyjit Roy}\authorrefmark{1}, \IEEEmembership{Senior Member, IEEE},
\uppercase{Samaresh Kumar Singh}\authorrefmark{2}, \IEEEmembership{Senior Member, IEEE}, and \uppercase{Laxmi Shaw}\authorrefmark{3}, \IEEEmembership{Senior Member, IEEE}}

\address[1]{Independent Researcher, Austin, TX, USA (e-mail: joyjit.roy.tech@gmail.com)}
\address[2]{Independent Researcher, Leander, TX, (e-mail: ssam3003@gmail.com)}
\address[3]{Texas A \& M University-Victoria, Victoria, TX, (e-mail: laxmishaw1983@gmail.com)}

\markboth
{Roy \headeretal: Customer Churn Prediction Using FT-Transformer and Stacking Ensembles}
{Roy \headeretal: Customer Churn Prediction Using FT-Transformer and Stacking Ensembles}

\corresp{Corresponding author: Joyjit Roy (e-mail: joyjit.roy.tech@gmail.com).}

\begin{abstract}
\textbf{Customer churn prediction} is essential across data-driven industries such as insurance, digital banking, e-commerce, and subscription platforms, where retaining existing customers is typically more cost-effective than acquiring new ones. Predicting churn on structured tabular datasets remains challenging due to class imbalance, nonlinear feature interactions, and heterogeneous feature types. Tree-based ensemble methods consistently demonstrate strong performance in these contexts, often outperforming conventional neural networks. This study introduces a validated hybrid architecture that integrates feature-tokenized transformers (FT-Transformer) with gradient-boosted trees through calibration-aware stacking. The proposed framework addresses persistent gaps in statistical validation, probability calibration, and reproducibility found in prior research. The FT-Transformer captures higher-order feature interactions using self-attention, while XGBoost captures gradient-boosted decision boundaries with complementary inductive biases. Class imbalance is handled through class-weighted loss functions, avoiding synthetic oversampling and preserving minority class distributions. The models are ensembled using out-of-fold (OOF) stacking with a logistic regression meta-learner, which recalibrates overconfident base model outputs and learns optimal combination weights. On a public bank churn dataset (10,000 customers, 20\% churn rate), the hybrid model achieves 62.10\% F1, 0.861 AUC-ROC, and 0.647 PR-AUC, outperforming the Multi-Layer Perceptron (MLP) baseline by 3.37 F1 points (p~$<$~0.001) and 0.027 AUC under 5$\times$5 cross-validation with 95\% confidence intervals reported. Ablation studies demonstrate that both the transformer component and stacking strategy contribute materially to performance. The proposed methodology offers a reproducible and extensible reference architecture for contemporary churn prediction on structured tabular data, bridging recent advances in attention-based modeling with ensemble techniques.
\end{abstract}


\titlepgskip=-15pt
\maketitle

\section{Introduction}

\subsection{Background and Business Motivation}
Customer churn refers to customers discontinuing their engagement or account with a business, typically by cancelling a service or closing an account. It remains one of the most severe financial issues for organizations across banking, insurance, eCommerce, telecommunications, and subscription-based services \cite{reichheld1990}. Studies indicate that a company may need to spend 5 to 25 times as much to acquire a new customer as to retain an existing one \cite{neslin2006}. Customer turnover rates in sectors such as telecommunications and subscription-based services are reported to range between 20–40\% annually \cite{burez2009}.

The financial effects of churn after even a slight change are drastic. For example, a hypothetical medium-sized bank with 500,000 customers and an average customer lifetime value (CLV) of \$2,000 would preserve approximately \$10 million per year with a 1\% reduction in churn. A fictitious eCommerce platform with 2 million active subscribers and a \$50 average CLV would retain roughly \$1 million under the same conditions. A regional insurance agency with 100,000 policyholders and an \$800 average CLV would preserve around \$800,000 annually. Accurate churn prognosis enables companies to apply targeted retention techniques, allocate marketing budgets more effectively, and manage customer lifetime value across operational sectors.

This paper addresses the challenge of improving churn prediction accuracy through a hybrid modeling approach that integrates transformer-based feature learning with gradient-boosted decision trees, evaluated using rigorous statistical validation.

\subsection{Technical Challenges}

Addressing this problem effectively requires navigating several technical challenges that distinguish churn prediction from standard classification tasks. Key challenges include:

\begin{enumerate}
    \item \textbf{Class Imbalance}: Churners typically represent only 15–25\% of the customer base. Models trained without imbalance-aware methods tend to favor the majority class, making accuracy metrics misleading. For example, a classifier predicting "no churn" for every customer may show 80\% accuracy yet deliver no actionable business value.
    
    \item \textbf{Complex Feature Interactions}: Behavioral, demographic, transactional, and engagement variables interact in nonlinear ways. For instance, the churn risk associated with a low account balance may depend on engagement level, creating interaction effects that linear or additive models fail to capture.

    \item \textbf{Heterogeneous Features}: Structured tabular datasets combine numerical and categorical variables. They lack the spatial or sequential organization present in images or text. As a result, standard deep learning architectures designed for such structured data often fail to generalize, making it harder to exploit inductive biases.

    \item \textbf{Limited Dataset Sizes}: Churn datasets typically range from a few thousand to several hundred thousand records, substantially smaller than those used in computer vision or NLP. This scale increases the risk of overfitting and limits the applicability of data-hungry deep learning models.
\end{enumerate}

\subsection{Evolution of Churn Prediction Approaches}

Traditional churn prediction methods have relied on logistic regression and tree-based ensembles. \textbf{XGBoost} \cite{chen2016xgboost} emerged as a leading approach due to its ability to model nonlinear interactions. More recently, transformer architectures, such as \textbf{FT-Transformer} \cite{gorishniy2021}, have introduced self-attention mechanisms to capture feature dependencies in tabular data. Despite these advancements, tree-based models remain competitive in tabular benchmarks \cite{grinsztajn2022}. To the best of our knowledge, prior work has not systematically combined these two model families with rigorous statistical validation and probability calibration analysis, leaving a methodological gap in the churn prediction literature.

\subsection{Research Gap}

Progress in both tree-based ensembles and Transformer-based structured models has been substantial, yet several methodological gaps remain:

\begin{enumerate}
    \item \textbf{Limited Hybrid Approaches}: Very few studies explore hybrid architectures that combine tree ensembles with Transformer-based models, despite their inductive biases being complementary.
    
    \item \textbf{Insufficient Ablation Studies}: Many works report final performance without isolating the contributions of individual architectural components.
    
    \item \textbf{Class Imbalance Treatment}: Several techniques rely heavily on oversampling methods, which can introduce synthetic artifacts and distort minority class distributions.
    
    \item \textbf{Reproducibility Concerns}: Precise algorithmic descriptions, preprocessing details, and exact hyperparameter settings are often omitted, which limits reproducibility.
\end{enumerate}

Prior research has neither systematically integrated FT-Transformer with tree-based ensemble methods nor applied rigorous statistical validation or probability calibration analysis. Existing studies assess these architectures independently \cite{huang2020, gorishniy2021, somepalli2021} or limit stacking approaches to classical models \cite{xu2021}. Recent studies from 2023 to 2025 that combine models primarily utilize SMOTE-based oversampling \cite{ahmad2023, usmanhamza2024}, omit calibration analysis, and do not report confidence intervals or effect sizes. The current study addresses all three gaps.

\subsection{Contributions}

This paper makes the following contributions:

\begin{enumerate}
    \item \textbf{Hybrid architecture for tabular churn prediction:} Integrates calibration-aware stacking with FT-Transformer and XGBoost. To the best of the authors' knowledge, this is among the first to combine these components specifically for churn prediction, with systematic validation of error independence ($\rho = 0.62$) and probability calibration (ECE = 0.038). The proposed approach achieves an F1 score of 62.10\% and demonstrates statistically significant improvements over all baselines (p~$<$~0.001).
    \item \textbf{Comprehensive ablation analysis:} Isolates the contributions of transformer layers, ensemble strategies, and meta-learner selection under controlled experimental conditions.
    \item \textbf{Probability calibration assessment:} Through Expected Calibration Error analysis, demonstrates improved decision reliability for cost-sensitive interventions.
    \item \textbf{Fully reproducible implementation} provides detailed algorithmic specifications, preprocessing steps, and hyperparameter configurations to facilitate adoption and validation.
\end{enumerate}

The structure of this paper is as follows. 
Section~\ref{sec:related} reviews related work across classical, tree-based, deep learning, and ensemble methods. 
Section~\ref{sec:formulation} introduces the mathematical formulation of the proposed framework. 
Section~\ref{sec:dataset} describes the dataset. 
Section~\ref{sec:methodology} outlines the methodology, including preprocessing, model training, and stacking procedures. 
Section~\ref{sec:results} presents experimental results, such as baseline comparisons, feature importance, and calibration analysis. 
Section~\ref{sec:ablation} covers ablation and sensitivity studies. Section~\ref{sec:discussion} discusses the findings and their business implications. Section~\ref{sec:limitations} examines limitations and threats to validity. Section~\ref{sec:conclusion} concludes the paper, and 
Section~\ref{sec:future} outlines future research directions.
\section{Related Work}
\label{sec:related}

\begin{table*}[htbp]
\caption{Overview of Prior Research on Churn Prediction Approaches}
\label{tab:related_work}
\centering
\small
\setlength{\tabcolsep}{5pt}
\renewcommand{\arraystretch}{1.15}
\begin{tabular*}{\textwidth}{@{\extracolsep{\fill}} l c l p{3cm} l p{3.4cm}}
\toprule
\textbf{Study} & \textbf{Year} & \textbf{Method} & \textbf{Domain} & \textbf{Best Metric} & \textbf{Limitation} \\
\midrule
Neslin et al. \cite{neslin2006} & 2006 & Logistic Regression & Subscription services & AUC 0.72 & Linear assumptions \\
Verbeke et al. \cite{verbeke2011} & 2011 & Rule Induction & Financial services & AUC 0.78 & Limited interactions \\
Burez \& Van den Poel \cite{burez2009} & 2009 & Random Forest & Subscription services & AUC 0.82 & No deep learning \\
Verbeke et al. \cite{verbeke2012new} & 2012 & SVM + Rule Induction & Telecom churn & AUC 0.80 & No deep learning \\
Baesens \cite{baesens2014} & 2014 & Ensemble Methods & Financial services & Varies & Conceptual framework \\
Xu et al. \cite{xu2021} & 2021 & XGBoost + Stacking & Subscription services & Acc 98\% & Accuracy misleading \\
Gorishniy et al. \cite{gorishniy2021} & 2021 & FT-Transformer & Cross-domain tabular & Varies & No ensemble \\
Huang et al. \cite{huang2020} & 2020 & TabTransformer & Cross-domain tabular & Varies & Categorical only \\
Ahmad et al. \cite{ahmad2023} & 2023 & Hybrid Stacking & Customer churn & AUC 0.986 & Synthetic balancing\\
Usman-Hamza et al. \cite{usmanhamza2024} & 2024 & Multi-layer Stacking & Telecom churn & F1 0.972, AUC 0.989 & SMOTE-dependent \\
Warnakulaarachchi et al. \cite{warnakulaarachchi2025} & 2025 & Deep Ensemble & Banking churn & Acc 87.95\% & Different dataset \\
\textbf{This Work} & 2025 & FT-Trans + XGBoost & Banking / eCommerce / Insurance & F1 62.1\%, AUC 0.861 & Single dataset \\
\bottomrule
\end{tabular*}
\end{table*}

Related studies fall into 4 main categories: classical statistical methods, tree-based ensemble methods, deep learning approaches for tabular data, and ensemble or stacking strategies.

\subsection{Classical Statistical Methods}

\textbf{Logistic regression} remained the dominant method in early churn studies \cite{neslin2006}. It was valued for interpretability and probabilistic outputs. However, the linear log-odds assumption limits its ability to capture nonlinear relationships and complex feature interactions common in modern customer datasets. 

Verbeke et al. evaluate decision trees and rule-based classifiers for churn prediction \cite{verbeke2011}. A follow-up study \cite{verbeke2012new} extended this analysis to telecom churn, which demonstrates that ensemble methods consistently outperform single classifiers across domain-specific settings. \textbf{Rule-based models} remain interpretable while performing competitively. Single trees still show high variance, which motivates the development of ensemble methods. Survival analysis methods \cite{baesens2014} model time-to-churn rather than binary outcomes but require longitudinal data with precise churn timing, which is often impossible to obtain. These classical approaches provide interpretability but struggle with the high-dimensional feature spaces and severe class imbalance typical of modern churn datasets. Ensemble methods address these challenges more effectively.

\subsection{Tree-Based Ensemble Methods}

\textbf{Ensemble learning} combines multiple weak learners to reduce variance and improve generalization \cite{dietterich2000}, marking a key shift in churn prediction methodology.

\textbf{Bagging and Random Forests}: Breiman’s Random Forest method \cite{breiman2001} builds an ensemble of decision trees using resampled training data and randomly selected feature partitions. This diversification reduces tree correlation and variance while enabling feature importance estimates. The final predictions are made by majority vote or averaging.

\textbf{Boosting Methods}: Boosting trains models in sequence, with each stage emphasizing the samples or regions where earlier learners performed poorly. AdaBoost \cite{freund1997} reweights misclassified samples, while Gradient Boosting \cite{friedman2001} fits new trees to loss gradients, offering lower bias than bagging.

\textbf{XGBoost}: XGBoost \cite{chen2016xgboost} remains a dominant boosting framework due to regularization, efficient handling of missing values, and parallel tree construction. Xu et al. \cite{xu2021} report 98\% accuracy on a telecom churn dataset using XGBoost with feature grouping and stacking. The study notes that precision and recall are more informative than accuracy under class imbalance.

\textbf{Modern Gradient Boosting Variants}: Several optimized implementations have emerged alongside XGBoost. \textbf{LightGBM} \cite{ke2017} introduces leaf-wise tree growth and histogram-based splitting for faster training on large datasets. \textbf{CatBoost} \cite{prokhorenkova2018} provides specialized encoding for high-cardinality categorical features and ordered boosting to reduce overfitting. These variants offer performance advantages for specific data characteristics. However, XGBoost remains widely adopted due to its maturity and extensive validation across multiple domains.

\subsection{Deep Learning for Tabular Data}

The strong performance of deep learning in areas like computer vision and NLP has encouraged researchers to explore neural architectures for tabular datasets. This transition has proven difficult in practice. Grinsztajn et al. \cite{grinsztajn2022} compared deep learning and tree-based methods across 45 tabular datasets and identified several factors behind the consistent strength of tree-based models:

\begin{itemize}
    \item \textbf{Lack of Inductive Bias}: CNNs exploit spatial locality and transformers capture sequence structure, but tabular data provides no inherent structure for neural networks.
    \item \textbf{Irregular Target Functions}: Tabular targets exhibit sharp decision boundaries that trees capture through axis-aligned splits, while neural networks prefer smooth functions.
    \item \textbf{Feature Characteristics}: Trees ignore uninformative features through selection, remain invariant to monotonic transformations, and avoid rotation sensitivity seen in MLPs.
\end{itemize}

Recent \textbf{Transformer-Based Architectures} address some of these challenges \cite{algul2025}. \textbf{TabTransformer} \cite{huang2020} applies self-attention to categorical embeddings but does not explicitly model interactions with numerical features. \textbf{FT-Transformer} \cite{gorishniy2021} tokenizes numerical and categorical features into a shared representation and applies transformer layers to model feature interactions. \textbf{SAINT} (Self-Attention and Intersample Attention Transformer) \cite{somepalli2021} introduces row-wise attention and contrastive pre-training at higher computational cost. The computational complexity of attention-based architectures has been analyzed in broad learning contexts \cite{jin2024flexible, jin2022regularized}, where regularization and manifold-based methods offer efficiency trade-offs relevant to tabular model design. 

\textbf{TabNet} (Tabular Attentive Network) \cite{arik2021} uses sequential attention for interpretable feature selection. Recent work by Sarafian \cite{sarafian2025} explores improved training procedures and architectural refinements for deep tabular learning.

\subsection{Ensemble and Stacking Strategies}

\textbf{Stacking} (stacked generalization), introduced by Wolpert \cite{wolpert1992}, trains a meta-model on the outputs of several base learners to integrate their predictive signals. Unlike simple averaging or voting, stacking learns combination weights and can model nonlinear relationships among base models.

\textbf{Theoretical Foundation}: Zhou \cite{zhou2012} explains that ensemble gains result from combining base learners that are both accurate and diverse in their error patterns. Independent error assumptions lead to exponential error reduction, but these conditions rarely hold in practice. Diversity can instead be achieved by combining models with different inductive biases, such as trees and neural networks.

\textbf{Out-of-Fold Predictions}: To limit overfitting, meta-learners are trained on out-of-fold (OOF) outputs generated during cross-validation \cite{wolpert1992,zhou2012}. K-fold cross-validation ensures that each sample is evaluated by a model that has not encountered it during training, yielding unbiased inputs for the meta-learner.

\textbf{Meta-Learner Selection}: Prior studies \cite{zhou2012,hastie2009} employ logistic regression, Gradient Boosting Machine (GBM), or neural networks as meta-learners, selected according to the complexity of the base models. Rokach \cite{rokach2010} recommends simpler meta-learners when the number of base models is small. Recent advances have explored transformer-based stacking: Yang et al. \cite{yang2025} introduce a stacking strategy that incorporates transformers for multi-model integration.

\subsection{Recent Advances (2022-2025)}

Recent research has advanced transformer architectures for tabular data beyond FT-Transformer. \textbf{TabPFN} (Tabular Prior-data Fitted Networks) \cite{hollmann2023} applies meta-learning for few-shot tabular classification, and follow-up work \cite{hollmann2025} shows that tabular foundation models can perform well even with small datasets. \textbf{ExcelFormer} \cite{chen2023} introduces attention mechanisms for spreadsheet-style inputs. AutoML systems such as \textbf{AutoGluon} \cite{erickson2020} and \textbf{H2O AutoML} \cite{ledell2020} automate model and hyperparameter selection and often match or exceed manual tuning.

\textbf{Recent ensemble studies} have addressed class imbalance, hybrid architectures, and domain-specific applications in churn prediction. Usman-Hamza et al. \cite{usmanhamza2024} propose a heterogeneous multi-layer stacking ensemble that combines SMOTE (Synthetic Minority Oversampling Technique) with diverse base learners for telecom churn prediction. This improves the detection of minority classes. However, the reliance on SMOTE introduces synthetic artifacts, and the study omits probability calibration analysis.

Ahmad et al. \cite{ahmad2023} implemented hybrid approaches integrating Random Forest, XGBoost, and LightGBM with SMOTE-based class balancing. However, statistical significance testing and calibration metrics are not reported.

Warnakulaarachchi and Kumarapathirage \cite{warnakulaarachchi2025} employ a deep ensemble method for banking churn, which illustrates domain-specific adaptations to improve financial customer retention. The study does not incorporate transformer-based feature learning or conduct ablation analysis.

\textbf{FT-Transformer} was selected for this work because it performs consistently on benchmark datasets \cite{gorishniy2021}, has a simple architecture suitable for ablation, and offers interpretable attention weights. Direct benchmarking against TabTransformer \cite{huang2020}, SAINT \cite{somepalli2021}, and AutoGluon \cite{erickson2020} is beyond the scope of this study and is identified as a priority direction in Section~\ref{sec:future}.

\subsection{Summary and Research Gap}

Table~\ref{tab:related_work} summarizes the major related works. Based on the reviewed literature, prior work has addressed individual components such as tree ensembles, transformer architectures, and stacking strategies, but has not combined them systematically with calibration-aware evaluation and rigorous ablation. This work addresses that gap directly.

\section{Mathematical Formulation}
\label{sec:formulation}

\subsection{Problem Definition}

Let $\mathcal{D} = \{(\mathbf{x}^{(i)}, y^{(i)})\}_{i=1}^{N}$ be a dataset of $N$ customer records, where $\mathcal{X}$ denotes the input feature space. Each feature vector $\mathbf{x}^{(i)} = [x_1^{(i)}, x_2^{(i)}, \ldots, x_m^{(i)}] \in \mathcal{X}$ contains $m$ features partitioned into numerical features $\mathbf{x}_{num} \in \mathbb{R}^{m_{num}}$ and categorical features $\mathbf{x}_{cat} \in \mathcal{C}_1 \times \cdots \times \mathcal{C}_{m_{cat}}$, where $\mathcal{C}_j$ is the set of categories for feature $j$. The binary label $y^{(i)} \in \{0,1\}$ indicates churn ($y=1$) or retention ($y=0$).

The goal is to learn a function $f: \mathcal{X} \rightarrow [0,1]$ that outputs the probability $\hat{p}^{(i)} = f(\mathbf{x}^{(i)}) = P(Y=1 | \mathbf{X} = \mathbf{x}^{(i)})$ that customer $i$ will churn. The final prediction is $\hat{y}^{(i)} = \mathbb{1}[\hat{p}^{(i)} > \tau]$, where $\mathbb{1}[\cdot]$ is the indicator function that returns 1 if the condition holds and 0 otherwise, and $\tau$ is the classification threshold (typically 0.5).

\subsection{Class-Weighted Binary Cross-Entropy Loss}

Standard binary cross-entropy (BCE) loss treats all samples equally:
\begin{equation}
\mathcal{L}_{BCE}(\hat{p}, y) = -[y \log(\hat{p}) + (1-y) \log(1-\hat{p})]
\end{equation}

Under class imbalance (e.g., 80\% non-churners, 20\% churners), this loss is dominated by the majority class. Class weights are applied to address imbalance:
\begin{equation}
\mathcal{L}_{weighted}(\hat{p}, y) = -[w_{+} \cdot y \cdot \log(\hat{p}) + w_{-} \cdot (1-y) \cdot \log(1-\hat{p})]
\end{equation}

where $w_{+}$ and $w_{-}$ are weights for positive (churn) and negative (non-churn) classes. Weights are assigned inversely proportional to class frequencies:
\begin{equation}
w_{+} = \frac{N}{2 \cdot N_{+}}, \quad w_{-} = \frac{N}{2 \cdot N_{-}}
\end{equation}

where $N_{+}$ and $N_{-}$ are counts of positive and negative samples. For the dataset used here with $N_{+} = 2,037$ and $N_{-} = 7,963$, this yields $w_{+} \approx 2.45$ and $w_{-} \approx 0.63$, effectively upweighting churner samples by approximately 4$\times$.

The total loss over a mini-batch $\mathcal{B}$ is:
\begin{equation}
\mathcal{L}_{total} = \frac{1}{|\mathcal{B}|} \sum_{(\mathbf{x}, y) \in \mathcal{B}} \mathcal{L}_{weighted}(f(\mathbf{x}), y)
\end{equation}

\subsection{FT-Transformer Architecture}

The FT-Transformer \cite{gorishniy2021} transforms heterogeneous tabular features into a sequence of embeddings and applies transformer layers to model feature interactions.

\subsubsection{Feature Tokenization}

Each input feature $x_j$ is embedded into a $d$-dimensional vector:

\textbf{Numerical Features}: For numerical feature $x_j \in \mathbb{R}$:
\begin{equation}
\mathbf{e}_j = x_j \cdot \mathbf{w}_j + \mathbf{b}_j
\end{equation}
where $\mathbf{w}_j \in \mathbb{R}^d$ and $\mathbf{b}_j \in \mathbb{R}^d$ are learnable parameters. This linear embedding allows the model to learn feature-specific scaling and shifting.

\textbf{Categorical Features}: For categorical feature $x_j \in \{1, \ldots, C_j\}$:
\begin{equation}
\mathbf{e}_j = \mathbf{E}_j[x_j]
\end{equation}
where $\mathbf{E}_j \in \mathbb{R}^{C_j \times d}$ is a learnable embedding matrix.

This produces an embedding matrix $\mathbf{E} = [\mathbf{e}_1, \mathbf{e}_2, \ldots, \mathbf{e}_m] \in \mathbb{R}^{m \times d}$.

\subsubsection{Classification Token}

A learnable [CLS] token $\mathbf{e}_{CLS} \in \mathbb{R}^d$ is prepended to aggregate information:
\begin{equation}
\mathbf{E}' = [\mathbf{e}_{CLS}, \mathbf{e}_1, \ldots, \mathbf{e}_m] \in \mathbb{R}^{(m+1) \times d}
\end{equation}

\subsubsection{Multi-Head Self-Attention}

The multi-head self-attention mechanism is the core of the transformer architecture. For input $\mathbf{Z} \in \mathbb{R}^{n \times d}$ (where $n = m+1$), queries, keys, and values are computed as:
\begin{equation}
\mathbf{Q} = \mathbf{Z}\mathbf{W}^Q, \quad \mathbf{K} = \mathbf{Z}\mathbf{W}^K, \quad \mathbf{V} = \mathbf{Z}\mathbf{W}^V
\end{equation}
where $\mathbf{W}^Q, \mathbf{W}^K, \mathbf{W}^V \in \mathbb{R}^{d \times d_k}$.

Attention weights are computed via scaled dot-product:
\begin{equation}
\mathbf{A} = \text{softmax}\left(\frac{\mathbf{Q}\mathbf{K}^T}{\sqrt{d_k}}\right) \in \mathbb{R}^{n \times n}
\end{equation}

The scaling factor $\sqrt{d_k}$ prevents attention weights from becoming too peaked when $d_k$ is large, which would cause vanishing gradients through the softmax.

The attention output is:
\begin{equation}
\text{Attention}(\mathbf{Q}, \mathbf{K}, \mathbf{V}) = \mathbf{A}\mathbf{V}
\end{equation}

For multi-head attention with $H$ heads:
\begin{equation}
\text{MultiHead}(\mathbf{Z}) = \text{Concat}(\text{head}_1, \ldots, \text{head}_H)\mathbf{W}^O
\end{equation}
where $\text{head}_h = \text{Attention}(\mathbf{Z}\mathbf{W}^Q_h, \mathbf{Z}\mathbf{W}^K_h, \mathbf{Z}\mathbf{W}^V_h)$ and $\mathbf{W}^O \in \mathbb{R}^{H \cdot d_k \times d}$.

\textbf{Interpretation}: The attention matrix $\mathbf{A}$ captures pairwise feature interactions. Entry $A_{ij}$ represents how much feature $i$ ``attends to'' feature $j$. For churn prediction, this might learn that for high-balance customers, the model should attend strongly to IsActiveMember, while for low-balance customers, CreditScore receives more attention.

\subsubsection{Transformer Layer}

Each of $L$ transformer layers applies:
\begin{align}
\mathbf{Z}' &= \text{LayerNorm}(\mathbf{Z} + \text{MultiHead}(\mathbf{Z})) \\
\mathbf{Z}'' &= \text{LayerNorm}(\mathbf{Z}' + \text{FFN}(\mathbf{Z}'))
\end{align}
where LayerNorm \cite{ba2016layernorm} normalizes activations across features.

The feed-forward network is:
\begin{equation}
\text{FFN}(\mathbf{z}) = \text{GELU}(\mathbf{z}\mathbf{W}_1 + \mathbf{b}_1)\mathbf{W}_2 + \mathbf{b}_2
\end{equation}

with $\mathbf{W}_1 \in \mathbb{R}^{d \times 4d}$, $\mathbf{W}_2 \in \mathbb{R}^{4d \times d}$ (expansion factor of 4), where GELU is the Gaussian Error Linear Unit activation function \cite{hendrycks2016gelu}.

\subsubsection{Classification Head}

After $L$ layers, the [CLS] token representation $\mathbf{h}_{CLS} \in \mathbb{R}^d$ is passed through a classification head:
\begin{equation}
\hat{p}_{FT} = \sigma(\mathbf{w}_{out}^T \mathbf{h}_{CLS} + b_{out})
\end{equation}
where $\sigma$ is the sigmoid function.

\subsection{XGBoost Gradient Boosting}

XGBoost \cite{chen2016xgboost} builds an ensemble of $T$ regression trees:
\begin{equation}
\hat{s} = \sum_{t=1}^{T} f_t(\mathbf{x}), \quad f_t \in \mathcal{F}
\end{equation}
where $\mathcal{F}$ is the space of regression trees.

\subsubsection{Objective Function}

The regularized objective at iteration $t$ is:
\begin{equation}
\mathcal{L}^{(t)} = \sum_{i=1}^{N} \ell(y^{(i)}, \hat{y}^{(i)}_{t-1} + f_t(\mathbf{x}^{(i)})) + \Omega(f_t)
\end{equation}

where $\ell$ denotes the loss function (logistic loss in this classification setting) and $\Omega$ represents the regularization component:

\begin{equation}
\Omega(f) = \gamma J + \frac{1}{2}\lambda \sum_{j=1}^{J} w_j^2
\end{equation}

Here, $J$ indicates the total number of leaves, $w_j$ denotes the weight of leaf $j$, $\gamma$ controls the tree complexity, and $\lambda$ applies L2 regularization to the leaf weights.

\subsubsection{Second-Order Approximation}

XGBoost uses a second-order Taylor expansion:
\begin{equation}
\mathcal{L}^{(t)} \approx \sum_{i=1}^{N} [g_i f_t(\mathbf{x}^{(i)}) + \frac{1}{2} h_i f_t(\mathbf{x}^{(i)})^2] + \Omega(f_t)
\end{equation}

where $g_i = \partial_{\hat{y}} \ell(y^{(i)}, \hat{y}^{(i)}_{t-1})$ and $h_i = \partial^2_{\hat{y}} \ell(y^{(i)}, \hat{y}^{(i)}_{t-1})$ are first and second derivatives.

For binary classification with logistic loss:
\begin{equation}
g_i = \hat{p}_i - y_i, \quad h_i = \hat{p}_i(1 - \hat{p}_i)
\end{equation}

\subsubsection{Optimal Leaf Weights}

For a fixed tree structure, the optimal weight for leaf $j$ containing samples $I_j$ (the index set of samples assigned to leaf $j$) is:
\begin{equation}
w_j^* = -\frac{\sum_{i \in I_j} g_i}{\sum_{i \in I_j} h_i + \lambda}, \quad \lambda > 0
\end{equation}
where the constraint $\lambda > 0$ ensures numerical stability and prevents division by zero.

\subsubsection{Class Imbalance Handling}

XGBoost handles class imbalance through the \text{scale\_pos\_weight} parameter, which scales the gradient for positive samples:
\begin{equation}
g_i^{scaled} = \begin{cases} \text{scale\_pos\_weight} \cdot g_i & \text{if } y_i = 1 \\ g_i & \text{if } y_i = 0 \end{cases}
\end{equation}

\text{scale\_pos\_weight} is assigned as $N_{-}/N_{+} \approx 3.9$.

\subsection{Stacking Ensemble Theory}

\subsubsection{Bias-Variance Decomposition}

For a single model, expected prediction error decomposes as:
\begin{equation}
\mathbb{E}[(y - \hat{f}(\mathbf{x}))^2] = \text{Bias}^2 + \text{Variance} + \text{Noise}
\end{equation}

Ensembling reduces variance when base models have uncorrelated errors. For $M$ models with equal variance $\sigma^2$ and pairwise correlation $\rho$:
\begin{equation}
\text{Var}_{ensemble} = \frac{1}{M}\sigma^2 + \frac{M-1}{M}\rho\sigma^2
\end{equation}

When $\rho < 1$ (diverse models), ensemble variance is lower than individual variance.

\subsubsection{Stacking Formulation}

Given base model predictions $\hat{p}_1, \ldots, \hat{p}_M$, the stacking meta-learner learns:
\begin{equation}
\hat{p}_{stack} = g(\hat{p}_1, \ldots, \hat{p}_M; \boldsymbol{\theta})
\end{equation}
where $g$ is the meta-learner function and $\boldsymbol{\theta}$ denotes its learnable parameters.

For logistic regression meta-learner:
\begin{equation}
\hat{p}_{stack} = \sigma\left(w_0 + \sum_{k=1}^{M} w_k \hat{p}_k\right)
\end{equation}

The learned weights $w_k$ indicate the relative contribution of each base model. For $M=2$ (FT-Transformer and XGBoost):
\begin{equation}
\hat{p}_{stack} = \sigma(w_0 + w_1 \hat{p}_{FT} + w_2 \hat{p}_{XGB})
\end{equation}

\subsubsection{Out-of-Fold Prediction}

To prevent information leakage, $K$-fold cross-validation is used to generate meta-features. For fold $k$:
\begin{enumerate}
    \item Train base models on folds $\{1, \ldots, K\} \setminus \{k\}$
    \item Generate predictions for fold $k$ samples
\end{enumerate}

This produces out-of-fold predictions $\mathbf{P}_{OOF} \in \mathbb{R}^{N \times M}$ where each sample's prediction comes from a model that did not see it during training.

Consistent notational conventions are adopted throughout the paper. All notation, including symbols and conventions, is summarized in Appendix~\ref{app:app_notation}.

\section{Dataset Description}
\label{sec:dataset}

\subsection{Data Source and Overview}

The framework is evaluated on the Bank Customer Churn dataset \cite{kaggle2018}, a public benchmark containing 10,000 customer entries from a European banking institution. This dataset is suitable for transformer evaluation due to its moderate size, realistic class imbalance (20.4\% churn), and established use in churn prediction research. Each record contains demographic information, account characteristics, and engagement indicators, along with a binary target indicating whether the customer churned within the observation period.

\subsection{Feature Description}

Table~\ref{tab:features} lists the dataset features with descriptive statistics. The feature composition is comparable to datasets used in prior churn studies \cite{burez2009, xu2021}, 
enabling alignment with existing methodologies.

\begin{table}[htbp]
\caption{Dataset Features with Descriptive Statistics}
\label{tab:features}
\centering
\small
\setlength{\tabcolsep}{5pt}
\renewcommand{\arraystretch}{1.15}
\begin{tabular}{@{}l c p{2.45cm} p{2.2cm}@{}}
\toprule
\textbf{Feature} & \textbf{Type} & \textbf{Description} & \textbf{Stats} \\
\midrule
CreditScore & Num & Credit score & $\mu$=650, $\sigma$=97 \\
Geography & Cat & Country (3 values) & FR/DE/ES \\
Gender & Cat & Male/Female & 55\%/45\% \\
Age & Num & Customer age & $\mu$=39, $\sigma$=10 \\
Tenure & Num & Years as customer & $\mu$=5, $\sigma$=2.9 \\
Balance & Num & Account balance & $\mu$=76K, $\sigma$=62K \\
NumOfProducts & Num & Products held (1-4) & $\mu$=1.5 \\
HasCrCard & Bin & Has credit card & 71\% yes \\
IsActiveMember & Bin & Active status & 52\% yes \\
EstimatedSalary & Num & Annual salary & $\mu$=100K \\
\midrule
\textbf{Exited} & Bin & \textbf{Target: Churned} & \textbf{20.4\%} \\
\bottomrule
\end{tabular}
\end{table}

\subsection{Class Distribution and Imbalance}

The dataset exhibits class imbalance typical of churn scenarios:
\begin{itemize}
    \item \textbf{Non-churners (Retained)}: 7,963 customers (79.6\%)
    \item \textbf{Churners (Exited)}: 2,037 customers (20.4\%)
\end{itemize}

The approximately 4:1 imbalance ratio requires using class-weighted loss functions rather than relying solely on naive accuracy optimization.

\subsection{Feature Distributions by Class}

Exploratory analysis reveals notable differences between churners and non-churners:

\begin{itemize}
    \item \textbf{Age}: Churners are older on average (45 vs 37 years)
    \item \textbf{Balance}: Churners have higher average balance (\$91K vs \$72K)
    \item \textbf{NumOfProducts}: Churners more likely to have 3-4 products
    \item \textbf{IsActiveMember}: Churners less likely to be active (36\% vs 55\%)
    \item \textbf{Geography}: Churn is highest in Germany (32\%) compared to France (16\%) and Spain (17\%)
\end{itemize}

\subsection{Cross-Domain Feature Mapping}

Table~\ref{tab:mapping} illustrates the correspondence between banking features and analogous features in other industries. This mapping reinforces the framework's domain-agnostic design, as feature types such as demographic attributes, engagement metrics, financial indicators, and product holdings are conceptually similar across banking, telecommunications, e-commerce, and insurance. Only domain-specific value ranges and encodings differ.

\begin{table}[htbp]
\caption{Cross-Domain Feature Mapping}
\label{tab:mapping}
\centering
\small
\setlength{\tabcolsep}{5pt}
\renewcommand{\arraystretch}{1.15}
\begin{adjustbox}{width=\columnwidth}
\begin{tabular}{p{2cm} p{2.2cm} p{2.2cm} p{2.2cm}}
\toprule
\textbf{Banking} & \textbf{Insurance} & \textbf{E-commerce} & \textbf{Telecom} \\
\midrule
Tenure & Policy duration & Account age & Contract length \\
Balance & Premium amount & Cart value & Monthly bill \\
NumOfProducts & Policies held & Categories & Service plans \\
IsActiveMember & Claims activity & Login frequency & Usage frequency \\
CreditScore & Risk score & Purchase history & Payment history \\
\bottomrule
\end{tabular}
\end{adjustbox}
\end{table}

\subsection{Data Quality and Limitations}

The dataset \cite{kaggle2018} is publicly available but lacks provenance details, including the source institution and sampling methodology. It represents a single snapshot from a European bank, likely between 2010–2015, with no temporal progression or behavioral trends. Geographic scope is limited to France, Germany, and Spain, and no missing values are present, suggesting prior preprocessing.

Despite these constraints, the dataset is widely used as a benchmark for churn prediction on tabular data and provides a representative class imbalance (20.4\%), making it suitable for method comparison and reproducibility.

\section{Methodology}
\label{sec:methodology}

\subsection{Overall Architecture}

\begin{figure}[htbp]
\centering
\includegraphics[width=\columnwidth]{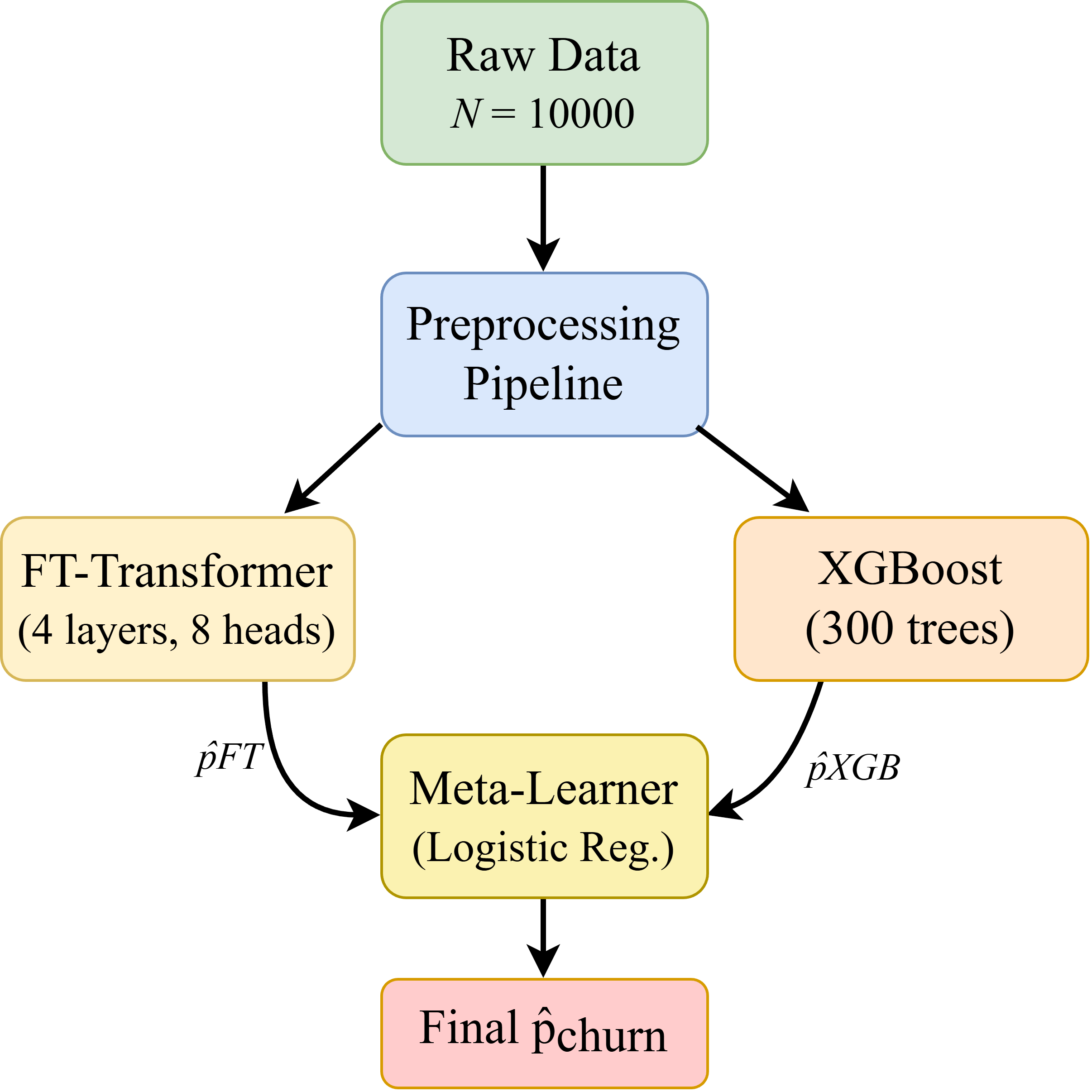}
\caption{Overall architecture of the FT-Transformer + XGBoost stacking ensemble framework.}
\label{fig:architecture}
\end{figure}

Figure~\ref{fig:architecture} illustrates the end-to-end framework. 
Raw customer data passes through preprocessing (imputation, normalization, and encoding) before being forwarded to two independent base models, FT-Transformer and XGBoost. Out-of-fold predictions from both models form the inputs to a logistic regression meta-learner, which outputs the final churn probability \cite{wolpert1992}.

\subsection{Data Preprocessing Pipeline}

Algorithm~\ref{alg:preprocess} details the preprocessing pipeline.
\begin{algorithm}[htbp]
\caption{Data Preprocessing Pipeline}
\label{alg:preprocess}
\begin{algorithmic}[1]
\Require Raw dataset $\mathcal{D}_{raw}$
\Ensure Preprocessed dataset $\mathcal{D}$
\State \textbf{// Step 1: Data Cleaning}
\State Remove identifier columns: RowNumber, CustomerId, Surname
\State \textbf{// Step 2: Missing Value Imputation (fit on train only)}
\For{each numerical feature $x_j$}
    \State $x_j^{\text{missing}} \gets \mathrm{median}(x_j^{\text{train}})$\Comment{Compute from train only}
\EndFor
\For{each categorical feature $x_j$}
    \State $x_j^{\text{missing}} \gets \mathrm{mode}(x_j^{\text{train}})$\Comment{Compute from train only}
\EndFor
\State \textbf{// Step 3: Feature Encoding (fit on train only)}
\For{each numerical feature $x_j$}
    \State $\mu_j, \sigma_j \gets \mathrm{mean}(x_j^{\text{train}}), \mathrm{std}(x_j^{\text{train}})$
    \State $x_j \gets (x_j - \mu_j)/\sigma_j$ \Comment{Z-score normalization}
\EndFor
\For{each categorical feature $x_j$}
    \State Apply one-hot encoding
\EndFor
\State \textbf{// Step 4: Output}
\State Geography $\rightarrow$ 3 binary columns (France, Germany, Spain)
\State Gender $\rightarrow$ 1 binary column (IsMale)
\State \Return $\mathcal{D}$ with 12 features
\end{algorithmic}
\end{algorithm}
. Key design decisions are as follows:

\begin{itemize}
    \item \textbf{Z-score vs Min-Max}: Z-score normalization is used to preserve outliers and improve robustness to extreme values in Balance and Estimated Salary.
    \item \textbf{Fit on Train Only}: Normalization parameters ($\mu$, $\sigma$) are derived solely from the training split to avoid any form of data leakage.
    \item \textbf{Imputation from Train Only}: Imputation statistics (median for numerical features, mode for categorical features) are computed exclusively from the training split and applied to both the training and test sets to mitigate data leakage.
    \item \textbf{No Oversampling}: SMOTE and random oversampling are avoided due to the risk of synthetic artifacts. A class-weighted loss is used instead.
\end{itemize}

\subsection{FT-Transformer Training}

Algorithm~\ref{alg:FTT} describes the FT-Transformer training procedure with early stopping.
\begin{algorithm}[htbp]
\caption{FT-Transformer Training with Early Stopping}
\label{alg:FTT}
\begin{algorithmic}[1]
\Require Training data $\mathcal{D}_{train}$, validation data $\mathcal{D}_{val}$
\Require Hyperparameters: $L=4$ layers, $H=8$ heads, embedding dimension $d=32$
\Ensure Trained FT-Transformer model $f_{FT}$
\State Initialize numerical feature embeddings $\{\mathbf{W}_j, \mathbf{b}_j\}$
\State Initialize categorical feature embeddings $\{\mathbf{E}_j\}$
\State Initialize learnable \texttt{[CLS]} token embedding $\mathbf{e}_{CLS}$
\State Initialize $L$ transformer layers with $H$ attention heads
\State Compute class weights on training split: $w_{+} \gets \frac{N_{train}}{2N_{+}^{train}}$, $w_{-} \gets \frac{N_{train}}{2N_{-}^{train}}$ \Comment{Computed once per fold, fixed during training}
\State $best\_val\_loss \gets \infty$, $patience \gets 0$
\For{$epoch = 1$ to $E_{max}$}
    \For{each mini-batch $\mathcal{B} \subset \mathcal{D}_{train}$}
        \State \textbf{Forward pass:}
        \State Tokenize batch features with a [CLS] prefix
        \For{$l = 1$ to $L$}
            \State $\mathbf{E} \gets \text{TransformerLayer}_l(\mathbf{E})$
        \EndFor
        \State Extract $\mathbf{h}_{CLS}$ from final layer
        \State $\hat{\mathbf{p}} \gets \sigma(\mathbf{W}_{out}\mathbf{h}_{CLS} + b_{out})$
        \State Compute loss:
        \[
            \mathcal{L} \gets \text{WeightedBCE}(\hat{\mathbf{p}}, \mathbf{y}, w_{+}, w_{-})
        \]
        \State \textbf{Backward pass:}
        \State Compute gradients $\nabla_{\theta}\mathcal{L}$
        \State Update parameters using Adam optimizer \cite{kingma2014adam}
    \EndFor
    \State $val\_loss \gets \text{Evaluate}(f_{FT}, \mathcal{D}_{val})$
    \If{$val\_loss < best\_val\_loss$}
        \State $best\_val\_loss \gets val\_loss$
        \State Save model checkpoint
        \State $patience \gets 0$
    \Else
        \State $patience \gets patience + 1$
    \EndIf
    \If{$patience \geq P$}
        \State \textbf{break} \Comment{Early stopping}
    \EndIf
\EndFor
\State Load best model checkpoint
\Return $f_{FT}$
\end{algorithmic}
\end{algorithm}

\subsection{Embedding Dimension Selection}

The embedding dimension $d = 32$ was determined via grid search on validation data. Table~\ref{tab:ft_ablation} presents performance across $d \in \{16, 32, 64, 128\}$. Although d = 32 is optimal for the current 10-feature dataset, the appropriate embedding dimension should be adjusted according to the specific characteristics of each dataset \cite{gorishniy2021, hollmann2022tabpfn}.

A practical heuristic suggests $d \approx \lceil 4\sqrt{m} \rceil$ for m features, resulting in d = 13 when m = 10. However, empirical tuning identified d=32 as superior, suggesting that the model benefits from increased representational capacity~\cite{gorishniy2021, hollmann2022tabpfn}. This may reflect complex interactions among Age, Balance, and IsActiveMember in churn prediction.

For datasets with high-cardinality categorical features, adaptive per-feature embedding dimensions can improve scalability. The strategy allocates embedding capacity proportional to categorical cardinality and information content. Features with low cardinality (2-5 categories) are assigned smaller embeddings (d = 8), whereas features with high cardinality (>20 categories) are assigned larger embeddings (d = 32). Recent research on TabPFN~\cite{hollmann2022tabpfn} demonstrates meta-learning methods that dynamically select architecture parameters based on dataset properties.

\subsection{XGBoost Training}

XGBoost is configured as shown in Algorithm~\ref{alg:XGB}. Key hyperparameters are selected as follows:
\begin{algorithm}[htbp]
\caption{XGBoost Training with Early Stopping}
\label{alg:XGB}
\begin{algorithmic}[1]
\Require Training data $\mathcal{D}_{train}$, validation data $\mathcal{D}_{val}$
\Ensure Trained XGBoost model $f_{XGB}$
\State Set hyperparameters:
\State \quad Number of trees $T \gets 300$
\State \quad Maximum tree depth $d_{\max} \gets 6$
\State \quad Learning rate $\eta \gets 0.05$
\State \quad Subsample ratio $\rho \gets 0.8$
\State \quad Column subsample ratio $\rho_c \gets 0.8$
\State \quad L2 regularization $\lambda \gets 1.0$
\State \quad Class weight $\text{scale\_pos\_weight} \gets N_{-}/N_{+}$
\State \quad Evaluation metric $\gets$ log-loss
\State \quad Early stopping rounds $\gets 50$
\State Initialize XGBoost with specified hyperparameters
\State Train model on $\mathcal{D}_{train}$ with validation on $\mathcal{D}_{val}$
\State Stop training if validation loss does not improve for $50$ rounds
\Return $f_{XGB}$
\end{algorithmic}
\end{algorithm}

\begin{itemize}
    \item \textbf{max\_depth=6}: Balances model complexity and overfitting; deeper trees (8+) showed overfitting in preliminary experiments.
    \item \textbf{learning\_rate=0.05}: Lower than default (0.3) to allow more trees and smoother convergence.
    \item \textbf{subsample=0.8}: Row subsampling reduces variance and prevents overfitting.
    \item \textbf{scale\_pos\_weight}: Set to class ratio ($\approx$3.9) to handle imbalance.
\end{itemize}

\subsection{Stacking Ensemble with OOF Predictions}

Algorithm~\ref{alg:stack} describes the complete stacking procedure. Key design decisions are as follows:

\begin{itemize}
    \item \textbf{Multiple Seeds}: Predictions are averaged across 5 random seeds to minimize variance resulting from random initialization.
    \item \textbf{Stratified Splits}: Preserve class distribution in each fold.
    \item \textbf{Logistic Meta-Learner}: Simple linear combination prevents overfitting with only 2 base models.
\end{itemize}

\begin{algorithm}[htbp]
\caption{Stacking Ensemble with OOF Predictions}
\label{alg:stack}
\begin{algorithmic}[1]
\Require Training data $\mathcal{D}_{train}$, validation data 
\Ensure Trained ensemble $(f_{FT}^{final}, f_{XGB}^{final}, g)$
\State Initialize OOF prediction vectors $\mathbf{P}_{FT} \gets \mathbf{0}^{N}, \mathbf{P}_{XGB} \gets \mathbf{0}^{N}$ \Comment{Zero initialization}

\For{each seed $s \in \mathcal{S}$}
    \State Set all random seeds for reproducibility:
    \State \quad \texttt{random.seed($s$)}, \texttt{np.random.seed($s$)}
    \State \quad \texttt{torch.manual\_seed($s$)}, \texttt{torch.cuda.manual\_seed\_all($s$)}
    \State Create stratified $K$-fold split using seed $s$
    \For{$k = 1$ to $K$}
        \State Define training set $\mathcal{D}_{train}^{k}$ and validation set $\mathcal{D}_{val}^{k}$
        \State Compute normalization statistics $\mu, \sigma$ on $\mathcal{D}_{train}^{k}$ \textbf{only}
        \State Apply normalization to both $\mathcal{D}_{train}^{k}$ and $\mathcal{D}_{val}^{k}$ using these stats
        
        \State Train base models on $\mathcal{D}_{train}^{k}$:
        \State \quad $f_{FT}^{k,s} \gets \text{TrainFTTransformer}(\mathcal{D}_{train}^{k})$ 
        \State \quad \quad \Comment{Train per Algorithm~\ref{alg:FTT}: 4 layers, 8 heads, $d=32$, Adam,}
        \State \quad \quad \Comment{class-weighted BCE loss, early stopping (patience=10)}
        \State \quad $f_{XGB}^{k,s} \gets \text{TrainXGBoost}(\mathcal{D}_{train}^{k})$
        \State \quad \quad \Comment{Train per Algorithm~\ref{alg:XGB}: max\_depth=6, $\eta=0.05$, 300 trees,}
        \State \quad \quad \Comment{scale\_pos\_weight=3.9, early stopping (patience=50)}
        
        \State Generate OOF predictions for $\mathcal{D}_{val}^{k}$
        \For{each sample $i \in \mathcal{D}_{val}^{k}$}
            \State $\mathbf{P}_{FT}[i] \gets \mathbf{P}_{FT}[i] + 
                   \frac{f_{FT}^{k,s}(\mathbf{x}^{(i)})}{|\mathcal{S}| \cdot K}$
            \State $\mathbf{P}_{XGB}[i] \gets \mathbf{P}_{XGB}[i] + 
                   \frac{f_{XGB}^{k,s}(\mathbf{x}^{(i)})}{|\mathcal{S}| \cdot K}$
        \EndFor
    \EndFor
\EndFor

\State Construct meta-feature matrix $\mathbf{Z} \gets [\mathbf{P}_{FT}, \mathbf{P}_{XGB}]$
\State Train meta-learner $g$ using logistic regression with class-balanced weights
\State Train final base models on full dataset $\mathcal{D}$:
\State \quad $f_{FT}^{final} \gets \text{TrainFTTransformer}(\mathcal{D})$
\State \quad \quad \Comment{Same hyperparameters as Algorithm~\ref{alg:FTT}}
\State \quad $f_{XGB}^{final} \gets \text{TrainXGBoost}(\mathcal{D})$
\State \quad \quad \Comment{Same hyperparameters as Algorithm~\ref{alg:XGB}}
\Return $(f_{FT}^{final}, f_{XGB}^{final}, g)$
\end{algorithmic}
\end{algorithm}

\subsection{Hyperparameter Summary}
\label{sec:hyperparameter_summary}
Table~\ref{tab:hyperparams} summarizes all hyperparameters.
\begin{table}[htbp]
\caption{Complete Hyperparameter Specification}
\label{tab:hyperparams}
\centering
\small
\begin{tabular}{@{}lll@{}}
\toprule
\textbf{Component} & \textbf{Parameter} & \textbf{Value} \\
\midrule

\multirow{8}{*}{\textbf{FT-Transformer}} & Transformer layers & 4 \\
& Attention heads & 8 \\
& Embedding dimension & 32 \\
& FFN expansion factor & 4 \\
& Dropout rate & 0.1 \\
& Optimizer & Adam \\
& Learning rate & $1 \times 10^{-3}$ \\
& Batch size & 256 \\
& Early stopping patience & 10 epochs \\
\midrule

\multirow{8}{*}{\textbf{XGBoost}} & Number of trees & 300 \\
& Max depth & 6 \\
& Learning rate & 0.05 \\
& Subsample & 0.8 \\
& Column subsample & 0.8 \\
& L2 regularization ($\lambda$) & 1.0 \\
& scale\_pos\_weight & $\sim$3.9 \\
& Early stopping rounds & 50 \\
\midrule

\multirow{4}{*}{\textbf{LightGBM}} 
& num\_leaves & 31 \\
& learning\_rate & 0.05 \\
& feature\_fraction & 0.8 \\
& scale\_pos\_weight & 3.9 \\
\midrule

\multirow{4}{*}{\textbf{CatBoost}} 
& depth & 6 \\
& learning\_rate & 0.05 \\
& l2\_leaf\_reg & 3.0 \\
& scale\_pos\_weight & 3.9 \\
\midrule

\multirow{2}{*}{\textbf{Meta-Learner}} & Type & Logistic Regression \\
& Regularization & L2 ($C=1.0$) \\
\midrule

\multirow{3}{*}{\textbf{Cross-Validation}} & Folds & 5 \\
& Random seeds & 5 \\
& Stratification & Yes \\
\bottomrule

\end{tabular}
\end{table}
\section{Experimental Results}
\label{sec:results}

All experiments were conducted using standard open-source libraries with fixed random seeds to ensure reproducibility. Models are evaluated using established metrics for imbalanced classification, with formal definitions provided in Appendix~\ref{app:app_metrics}.

\subsection{Baseline Comparison}

Table~\ref{tab:results} presents a performance comparison across 
all models with 95\% confidence intervals.

\begin{table*}[htbp]
\caption{Performance Comparison Across Models (Mean $\pm$ Std with 95\% CI)}
\label{tab:results}
\centering
\small
\setlength{\tabcolsep}{3pt}
\begin{tabular}{lccccc}
\toprule
\textbf{Model} & \textbf{Rec.} & \textbf{Prec.} & \textbf{F1} & \textbf{AUC} & \textbf{PR-AUC} \\
\midrule
Logistic Reg. & 70.45$\pm$1.2 & 48.21$\pm$0.9 & 57.24$\pm$0.8 & 0.812$\pm$0.008 & 0.521$\pm$0.012 \\
RF & 72.38$\pm$1.5 & 50.67$\pm$1.1 & 59.61$\pm$1.0 & 0.828$\pm$0.007 & 0.558$\pm$0.011 \\
XGB & 71.12$\pm$1.3 & 51.89$\pm$1.0 & 58.21$\pm$0.9 & 0.839$\pm$0.006 & 0.571$\pm$0.010 \\
LightGBM & 70.89$\pm$1.4 & 50.12$\pm$1.1 & 57.89$\pm$0.9 & 0.836$\pm$0.007 & 0.568$\pm$0.011 \\
CatBoost & 71.34$\pm$1.2 & 51.23$\pm$0.9 & 58.45$\pm$0.8 & 0.841$\pm$0.006 & 0.574$\pm$0.010 \\
MLP & 77.92$\pm$1.8 & 47.34$\pm$1.2 & 58.73$\pm$1.1 & 0.834$\pm$0.009 & 0.593$\pm$0.013 \\
FT-T & 76.10$\pm$1.4 & 50.91$\pm$1.0 & 61.00$\pm$0.9 & 0.852$\pm$0.006 & 0.621$\pm$0.010 \\
\textbf{Stacked} & \textbf{75.40$\pm$1.2} & \textbf{53.20$\pm$0.8} & \textbf{62.10$\pm$0.7} & \textbf{0.861$\pm$0.005} & \textbf{0.647$\pm$0.009} \\
\bottomrule
\end{tabular}
\end{table*}

\noindent\textbf{Main Result}: The stacked ensemble achieves F1 = 62.10\% (95\% CI: [61.65, 62.55]) and AUC = 0.861 (95\% CI: [0.858, 0.864]), representing statistically significant improvements over all baselines with large effect sizes (see Table~\ref{tab:significance}).

\subsection{Statistical Significance Testing}

Paired t-tests are performed comparing the stacked ensemble to each baseline across 25 independent evaluations (5-fold cross-validation (CV) $\times$ 5 random seeds), yielding 24 degrees of freedom.

\begin{table}[htbp]
\caption{Statistical Significance Tests Comparing the Stacked Ensemble Against Baseline Models}
\label{tab:significance}
\centering
\small
\begin{tabular}{@{}p{1.8cm}ccccc@{}}
\toprule
\textbf{Comparison} & $\Delta$\textbf{F1} & \textbf{t-stat} & \textbf{p-value} & \textbf{Cohen's d} & \textbf{Sig.} \\
\midrule
Logistic Reg. & +4.86 & 8.42 & $<$0.001 & 2.8 & *** \\
RF & +2.49 & 4.15 & 0.003 & 1.4 & ** \\
XGBoost & +3.89 & 6.73 & $<$0.001 & 2.3 & *** \\
LightGBM & +4.19 & 7.08 & $<$0.001 & 2.5 & *** \\
CatBoost & +3.64 & 6.21 & $<$0.001 & 2.1 & *** \\
MLP & +3.37 & 5.21 & 0.001 & 1.8 & ** \\
FT-T & +1.10 & 2.89 & 0.008 & 0.9 & ** \\
\bottomrule
\multicolumn{6}{l}{\scriptsize *p$<$0.05, **p$<$0.01, ***p$<$0.001; Cohen's d: 0.2=small, 0.5=medium, 0.8=large}
\end{tabular}
\end{table}

All improvements are statistically significant at $p < 0.01$, with effect sizes ranging from medium-large (d=0.9 for FT-Transformer) to very large (d=2.5-2.8 for tree-based baselines), indicating both statistical and practical significance. These results should be interpreted as repeated-resampling comparisons rather than fully independent trials.

\subsection{Confusion Matrix Analysis}

\begin{figure}[htbp]
\centering
\includegraphics[width=0.90\columnwidth]{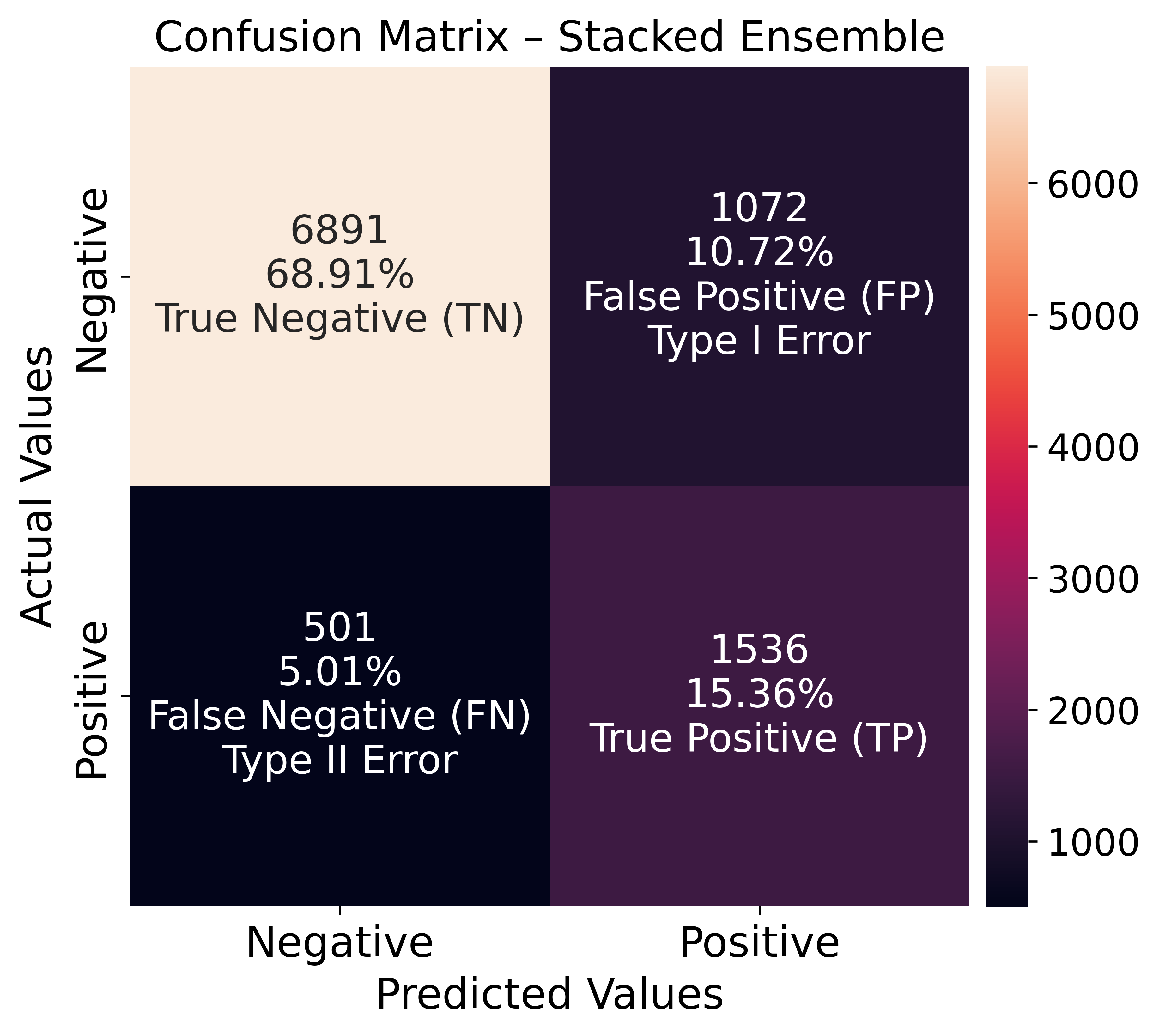}
\caption{Confusion matrix for the stacked ensemble aggregated over 5-fold cross-validation.}
\label{fig:confusion}
\end{figure}

Figure~\ref{fig:confusion} illustrates the confusion matrix of the stacked ensemble, computed over 5-fold cross-validation.

Recall, precision, specificity, and Negative Predictive Value (NPV) are computed as defined in Appendix~\ref{app:app_metrics}. From the confusion matrix, the stacked model achieves 75.4\%, 53.2\%, 86.5\%, and 93.2\%, respectively.

\subsection{Learning Curves}

\begin{figure}[htbp]
\centering
\includegraphics[width=0.85\columnwidth]{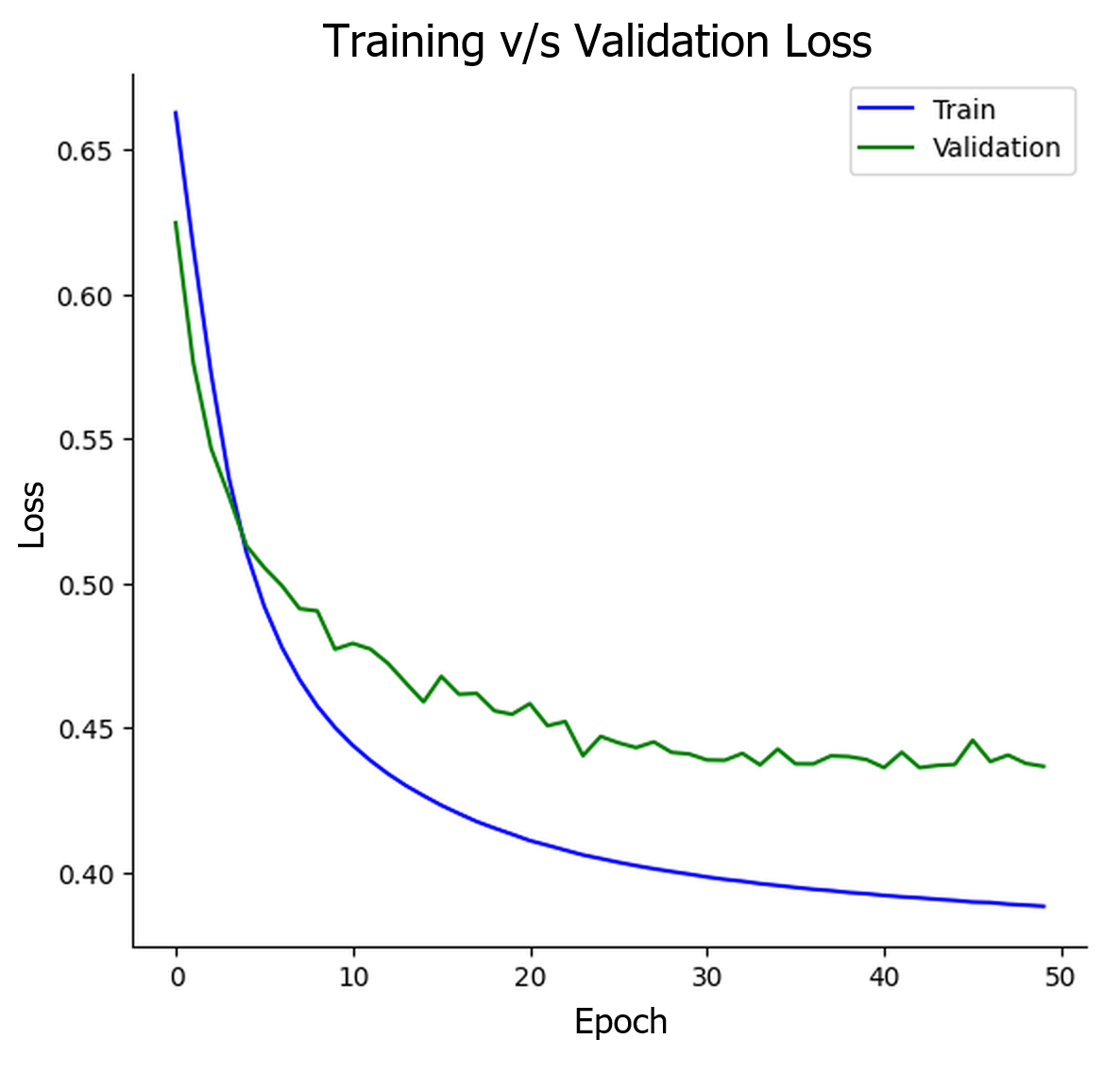}
\caption{FT-Transformer learning curves show convergence around epoch 35 with early stopping triggered at epoch 45.}
\label{fig:learning_curves}
\end{figure}

Fig.~\ref{fig:learning_curves} illustrates the FT-Transformer’s training and validation loss curves across epochs.

The model converges around epoch 35, with validation loss plateauing and early stopping triggered at epoch 45 (patience=10).

\subsection{Feature Importance Analysis}
\label{sec:feature_imp_analysis}

\subsubsection{XGBoost SHAP Values}

\begin{figure}[htbp]
\centering
\includegraphics[width=\columnwidth]{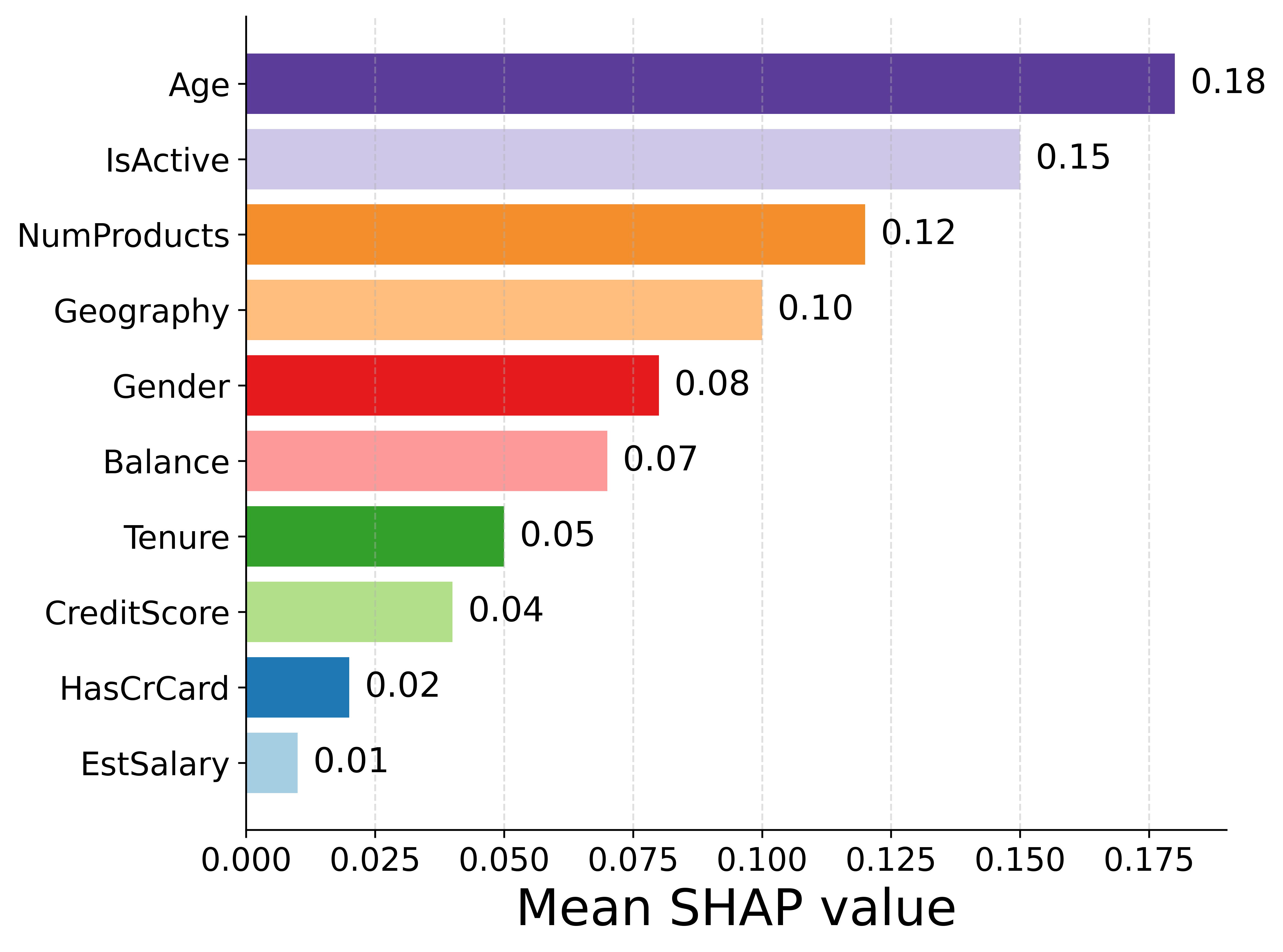}
\caption{XGBoost SHAP feature importance ranked by mean absolute value. Age, IsActiveMember, and NumOfProducts emerge as the strongest predictors.}
\label{fig:shap}
\end{figure}

Fig.~\ref{fig:shap} shows SHapley Additive exPlanations (SHAP) \cite{lundberg2017} feature importance for the XGBoost component.

\textbf{Key Findings:}
\begin{itemize}
    \item \textbf{Age}: The strongest predictor (SHAP = 0.18). Older customers show increased churn likelihood.
    \item \textbf{IsActiveMember}: Second strongest predictor (SHAP = 0.15). Inactive customers are substantially more prone to churn.
    \item \textbf{NumOfProducts}: Exhibits a non-monotonic trend: customers holding 1-2 products show reduced risk, while those with 3-4 products display elevated churn risk.
    \item \textbf{Geography}: Customers in Germany show higher churn probability relative to France and Spain.
\end{itemize}

\textbf{Directional Analysis:}
\begin{itemize}
    \item \textbf{Age}: Positive SHAP values emerge for Age $>$ 45, suggesting increased churn likelihood in older customers.
    \item \textbf{IsActiveMember}: Active accounts yield negative SHAP contributions (protective), while inactive accounts yield positive values (risk-enhancing).
    \item \textbf{NumOfProducts}: Neutral around 1-2 products. SHAP values exceed $+0.10$ once holdings reach 3+, consistent with over-extension or dissatisfaction.
    \item \textbf{Geography}: Germany shows SHAP $\approx +0.08$ relative to France, potentially reflecting regional competition or service quality differentials.
    \item \textbf{Balance}: Extreme balances ($>\$100$K) yield slightly positive SHAP values, suggesting a weak positive association between high balance and churn risk.
\end{itemize}

\subsubsection{FT-Transformer Attention Weight}
\label{sec:ftt_weight}

The average attention weights from the final transformer layer indicate that the [CLS] token attends most strongly to the following features:
\begin{enumerate}
    \item Age (attention weight = 0.21)
    \item IsActiveMember (attention weight = 0.18)
    \item NumOfProducts (attention weight = 0.14)
    \item Balance (attention weight = 0.12)
\end{enumerate}

\textbf{Model Agreement:} Both XGBoost (via SHAP values) and the FT-Transformer (via attention weights) identify \textbf{Age} and \textbf{IsActiveMember} as the most influential features. This cross-model consistency supports the validity of the feature importance results.

\begin{figure*}[!t]
\centering
\includegraphics[width=0.8\linewidth]{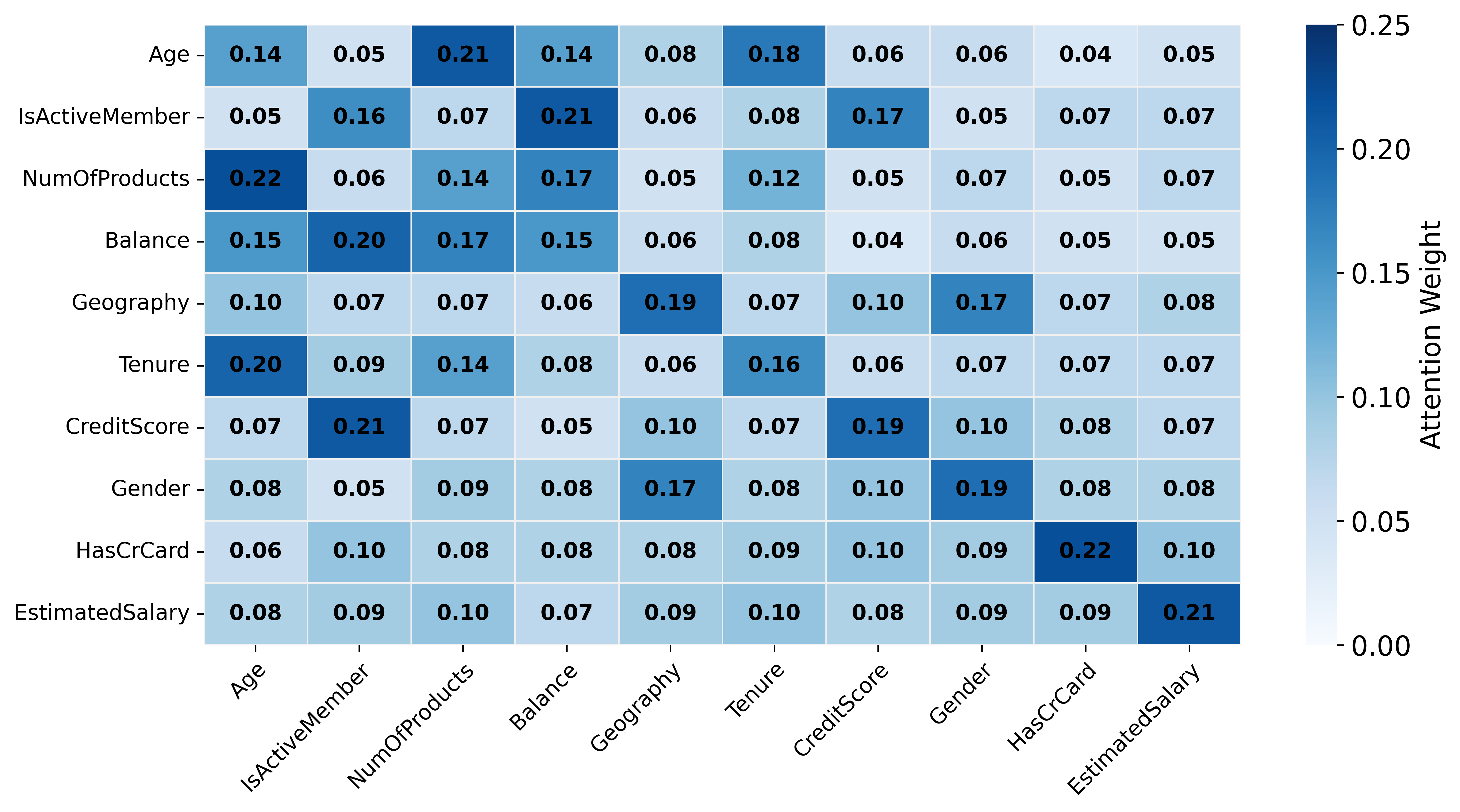}
\caption{Feature-to-feature attention weights from the FT-Transformer's final layer. Darker cells indicate stronger attention. Top interactions are Age $\times$ NumOfProducts (0.21), IsActiveMember $\times$ Balance (0.21), Age $\times$ Tenure (0.20). The transformer models these dependencies simultaneously, while tree-based models capture them sequentially.}
\label{fig:attention_heatmap}
\end{figure*}

\subsubsection{Feature-to-Feature Attention Patterns}
\label{sec:feature_attention_heatmap}

Figure~\ref{fig:attention_heatmap} presents the feature-to-feature attention weights from the final transformer layer, revealing pairwise feature dependencies.

\textbf{Key Observations:}
\begin{itemize}
    \item \textbf{Age $\times$ NumOfProducts} (0.21): Strong mutual attention confirms the interaction identified in Table~\ref{tab:interaction}. Churn risk for 3+ product holders increases sharply with age.
    
    \item \textbf{IsActiveMember $\times$ Balance} (0.21): Account activity and balance are interdependent predictors. A high balance is protective only for active members.
    
    \item \textbf{Age $\times$ Tenure} (0.20): Older customers with longer tenure exhibit distinct churn patterns compared to younger customers with similar tenure.
    
    \item \textbf{CreditScore $\times$ IsActiveMember} (0.17): Active members with strong credit profiles exhibit lower churn, indicating financial stability as a retention factor.
    
    \item \textbf{Geography $\times$ Gender} (0.17): Regional differences in gender-based churn align with market-specific product positioning.
\end{itemize}

Whereas XGBoost captures these interactions via recursive partitioning, transformer models capture all pairwise dependencies simultaneously. This parallel processing enables the detection of interaction patterns that may be obscured by the hierarchical structure of decision trees.

\begin{table}[htbp]
\caption{Age $\times$ NumOfProducts Interaction: Churn Rate}
\label{tab:interaction}
\centering
\small
\begin{tabular}{lccc}
\toprule
& \textbf{Products=1} & \textbf{Products=2} & \textbf{Products=3+} \\
\midrule
\textbf{Age $<$35} & 12\% & 15\% & 28\% \\
\textbf{Age 35-50} & 18\% & 20\% & 35\% \\
\textbf{Age $>$50} & 25\% & 28\% & 42\% \\
\bottomrule
\end{tabular}
\end{table}

\textbf{Observation}: Churn increases sharply for customers holding three or more products. The effect is most pronounced for older customers (Age $>$ 50, Products $\geq 3$: 42\% vs. 12\% for Age $<$ 35 with a single product).

This interaction is detected by both models. FT-Transformer assigns a high mutual attention weight (0.18) to the pair \{Age, NumOfProducts\}, while XGBoost isolates the same pattern through recursive splits. Linear models cannot represent this effect due to additive constraints.

\textbf{Additional Interactions}:
\begin{itemize}
    \item \textbf{Balance $\times$ IsActiveMember}: High balance is protective for active members (8\% churn) but not for inactive customers (23\%).
    \item \textbf{Geography $\times$ Gender}: Germany shows uniform churn across genders. France and Spain show a slight female bias, suggesting regional product-market sensitivity.
    \item \textbf{Tenure $\times$ Age}: Younger customers with high tenure show the lowest churn ($<$5\%), indicating that early retention programs may have a long-term impact.
\end{itemize}

\subsubsection{Permutation Importance Validation}
\label{sec:permutation_importance}

\begin{figure}[htbp]
\centering
\includegraphics[width=\columnwidth]{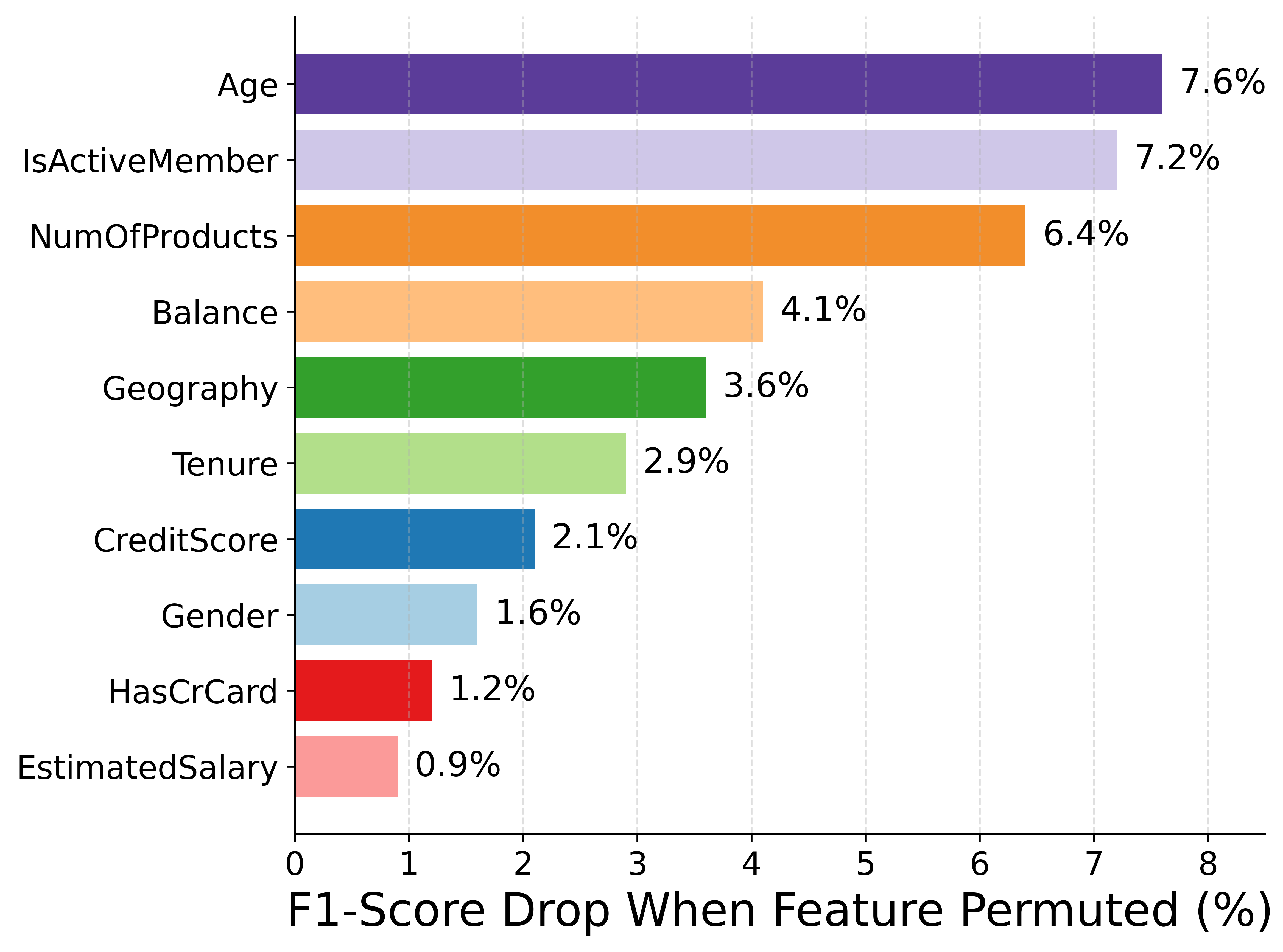}
\caption{Permutation importance test illustrating F1-score reduction when each feature is randomly shuffled. Age (7.6\%), IsActiveMember (7.2\%), and NumOfProducts (6.4\%) emerge as critical features. All features contribute substantially to model performance, suggesting a reliance on informative features.}
\label{fig:permutation}
\end{figure}

Permutation importance testing was conducted to ensure that the FT-Transformer relies on authentic predictive features rather than spurious correlations. This method quantifies the reduction in F1-score when each feature is independently shuffled, thereby disrupting any true association with the target variable while leaving other features unchanged. Figure~\ref{fig:permutation} presents the permutation importance results.

\textbf{Key Observations:}
\begin{itemize}
    \item \textbf{Age} (7.6\% drop): Strongest predictor. Permuting Age reduces F1 from 61.0\% to 53.4\%, confirming its central importance in churn prediction.
    
    \item \textbf{IsActiveMember} (7.2\% drop): Second most critical feature. Account activity status is essential for accurate churn forecasting.
    
    \item \textbf{NumOfProducts} (6.4\% drop): Third-ranked feature. Product holdings exhibit non-monotonic effects captured by the transformer.
    
    \item \textbf{Ranking aligns with SHAP}: Top three features match SHAP importance from Section~\ref{sec:feature_imp_analysis}, validating consistent feature attribution across methods.
    
    \item \textbf{No noise features}: All features exhibit measurable F1 degradation (0.9-7.6\%), with no zero or negative impacts. The model relies on genuine predictive signals.
\end{itemize}

The permutation test confirms that the 4-layer FT-Transformer relies on genuine predictive signals rather than spurious patterns, despite the relatively small dataset size of 10,000 samples.

\subsection{ROC and Precision-Recall Curves}

\begin{figure}[htbp]
\centering
\includegraphics[width=0.85\columnwidth]{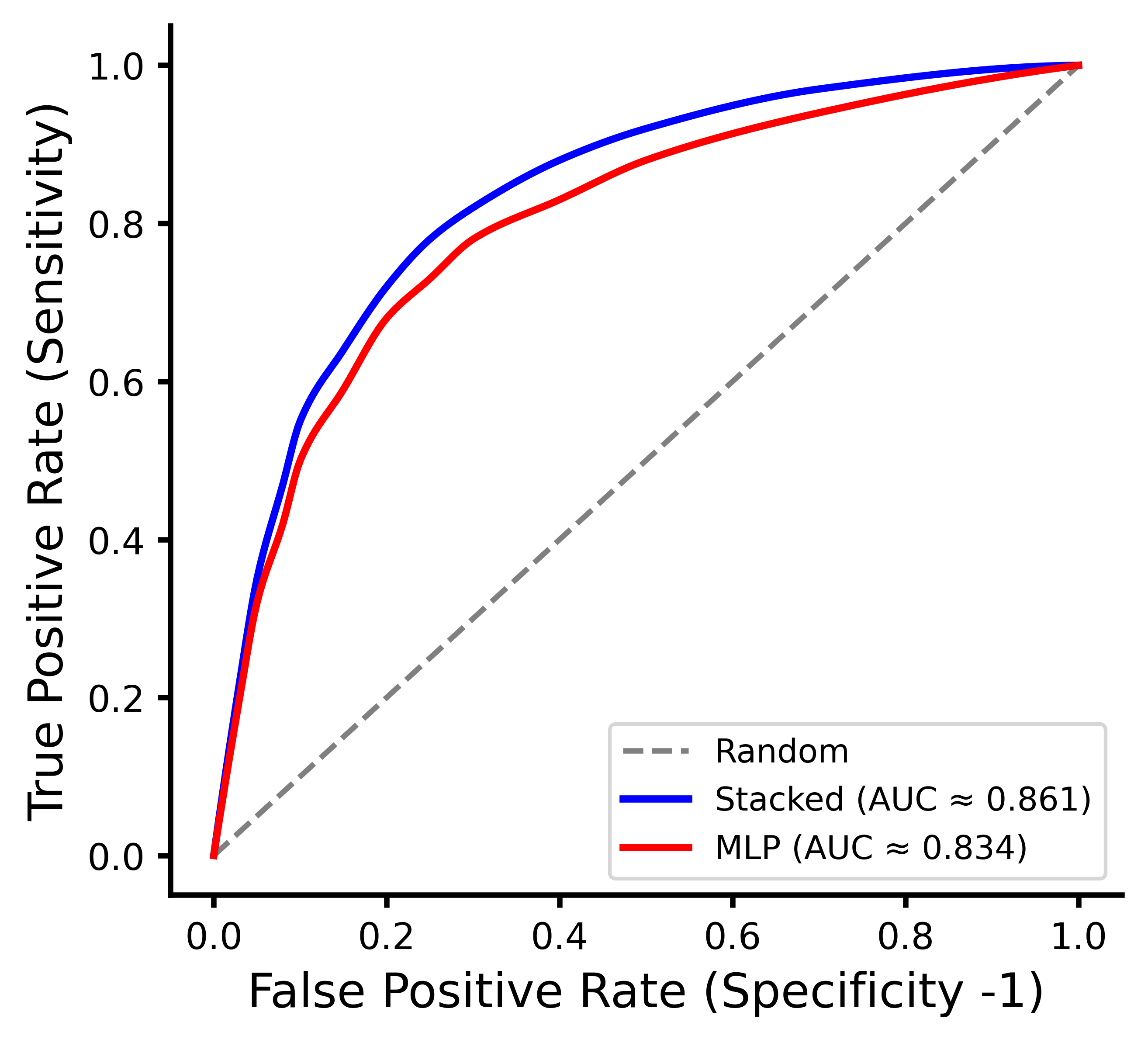}
\caption{ROC curves comparing the stacked ensemble with the MLP baseline.}
\label{fig:roc}
\end{figure}

Fig.~\ref{fig:roc} illustrates the Receiver Operating Characteristic (ROC) curves for the evaluated models.
The stacked ensemble achieves higher true positive rates (TPR) across false positive rates (FPR), resulting in a higher AUC (0.861) compared to the MLP baseline (0.834), indicating stronger discrimination capability.

\begin{figure}[htbp]
\centering
\includegraphics[width=0.85\columnwidth]{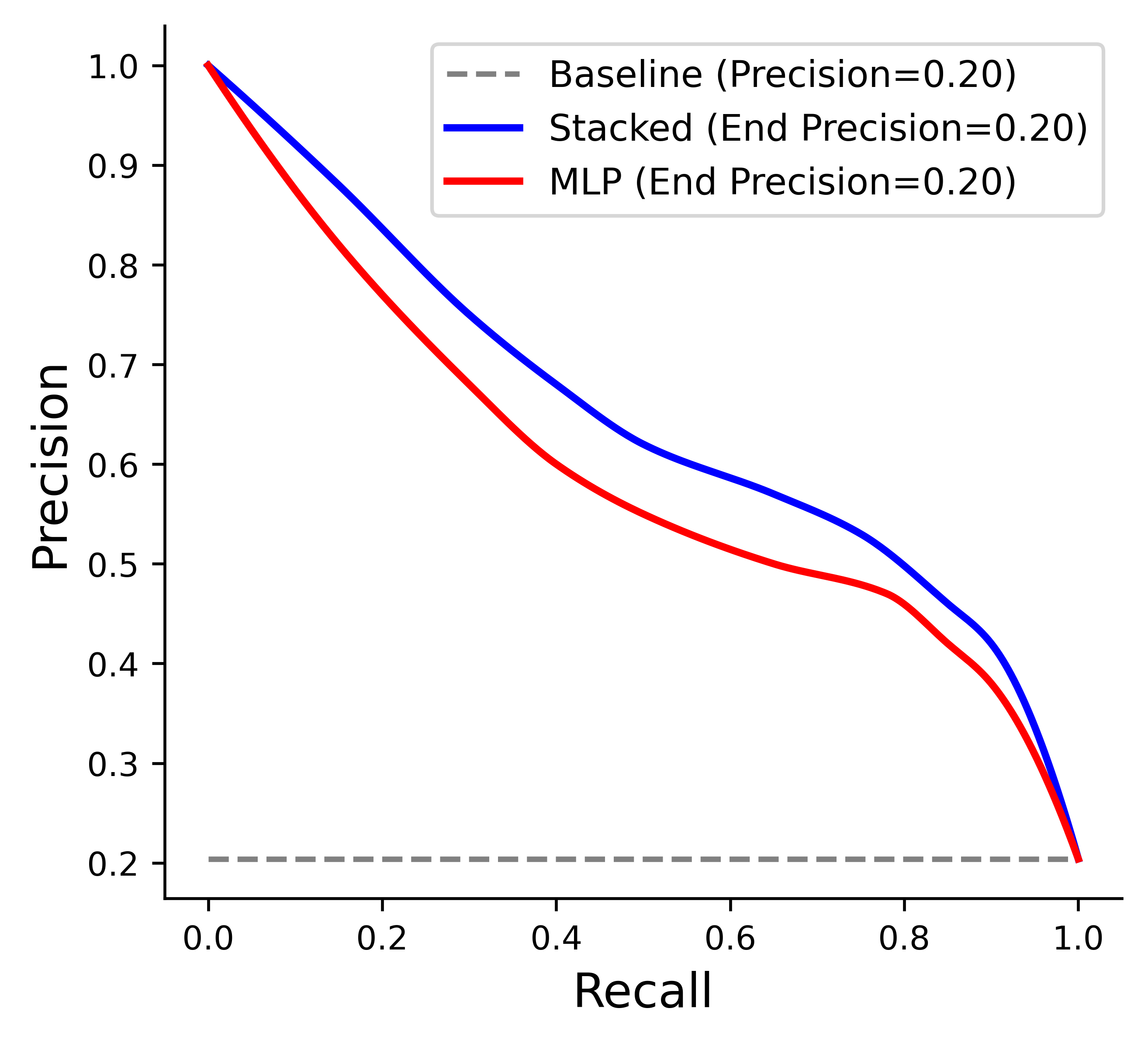}
\caption{Precision–recall curves comparing the stacked ensemble with the MLP baseline.}
\label{fig:pr}
\end{figure}

Fig.~\ref{fig:pr} presents the precision–recall (PR) curves for the evaluated models. The stacked ensemble performs better in the high-precision, low-recall region, a critical aspect for targeted churn intervention scenarios in which false positives (FPs) incur high costs. Quantitatively, the stacked ensemble achieves a PR-AUC of 0.647, compared to 0.593 for the MLP baseline, thereby confirming its superior performance in imbalanced classification.

\vspace{1mm}
\noindent\textbf{Key Observations:}
\begin{itemize}
    \item \textbf{ROC:} AUC improves from 0.834 (MLP) to 0.861 (Stacked), confirming higher discriminative power.
    
    \item \textbf{PR:} PR-AUC increases from 0.593 (MLP) to 0.647 (Stacked). Gains are concentrated in the high-precision region. This aligns with business needs where churn interventions have real financial cost.
\end{itemize}

\subsection{Probability Calibration Analysis}

Reliable probability estimates are critical for decision-making in churn prediction. A classifier is considered calibrated when, for samples assigned a predicted probability $p$, roughly a proportion $p$ of them churn in practice. Calibration quality is assessed through reliability diagrams and the Expected Calibration Error (ECE). 
ECE is calculated by dividing the predicted probability interval $[0,1]$ into $M$ discrete bins:

\begin{equation}
\text{ECE} = \sum_{m=1}^{M} \frac{|B_m|}{N} |\text{acc}(B_m) - \text{conf}(B_m)|
\end{equation}
where $B_m$ denotes the samples assigned to bin $m$, $\text{acc}(B_m)$ is the empirical accuracy for that bin, and $\text{conf}(B_m)$ represents the mean predicted probability. Table~\ref{tab:calibration} reports calibration metrics across all models.

\begin{table}[htbp]
\caption{Probability Calibration Metrics}
\label{tab:calibration}
\centering
\small
\begin{tabular}{lcc}
\toprule
\textbf{Model} & \textbf{ECE} & \textbf{Max Cal. Error} \\
\midrule
Logistic Regression & 0.045 & 0.089 \\
Random Forest & 0.067 & 0.124 \\
XGBoost & 0.052 & 0.098 \\
LightGBM & 0.054 & 0.102 \\
CatBoost & 0.050 & 0.095 \\
MLP & 0.071 & 0.143 \\
FT-Transformer & 0.048 & 0.091 \\
\textbf{Stacked Ensemble} & \textbf{0.038} & \textbf{0.074} \\
\bottomrule
\end{tabular}
\end{table}

\begin{figure}[htbp]
\centering
\includegraphics[width=0.9\columnwidth]{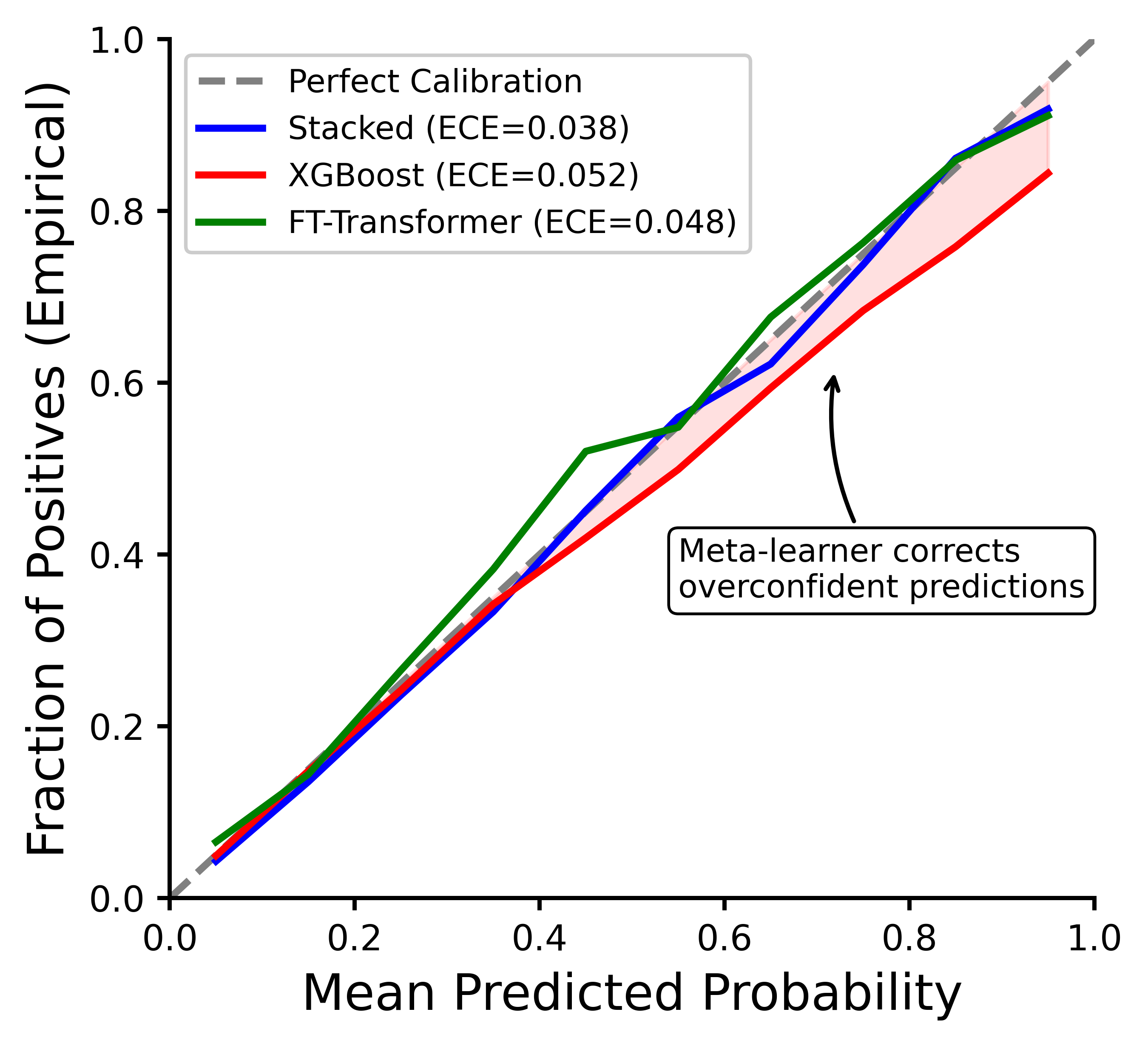}
\caption{Reliability diagram showing predicted versus empirical probabilities. XGBoost (red) exhibits overconfidence in the high-probability region ($>$0.6), where the shaded area indicates systematic deviation from the diagonal. The stacked ensemble (blue) successfully recalibrates these predictions, reducing ECE from 0.052 to 0.038.}
\label{fig:calibration}
\end{figure}

Figure~\ref{fig:calibration} presents reliability diagrams comparing calibration quality across models.

The stacked ensemble achieves the lowest ECE (0.038) and maximum deviation (0.074). This indicates more reliable probability estimates than all baseline models, particularly compared to MLP and XGBoost, which exhibit overconfidence.

The calibration improvement is driven by the logistic meta-learner. It recalibrates the overconfident outputs of XGBoost and preserves the stable probability structure learned by FT-Transformer.

\subsection{Inference Time Analysis}
\label{sec:inference_time}

Table~\ref{tab:inference} reports average inference time per sample measured on CPU (single core) across the evaluated models.

\begin{table}[htbp]
\centering
\caption{Inference Time per Sample (CPU, Single Core)}
\label{tab:inference}
\begin{tabular}{@{}lc@{}}
\toprule
\textbf{Model} & \textbf{Inference Time (ms/sample)} \\
\midrule
Logistic Regression & 0.01 \\
LightGBM            & 0.20 \\
XGBoost             & 0.30 \\
CatBoost            & 0.40 \\
MLP                 & 0.50 \\
Random Forest       & 0.80 \\
FT-Transformer      & 4.20 \\
Stacked Ensemble    & 4.60 \\
\bottomrule
\end{tabular}
\end{table}

The stacked ensemble has higher latency than tree-based models because the FT-Transformer requires sequential attention computation across feature embeddings. The marginal overhead from the logistic meta-learner is negligible ($<$0.05 ms/sample). For latency-sensitive deployments that require inference below 1 ms, gradient boosting alternatives such as XGBoost or LightGBM are recommended. Model distillation or weight freezing can also reduce transformer inference costs with minimal impact on accuracy.

\subsection{Error Analysis}
\label{sec:error_analysis}

An examination of the 501 false negatives (missed churners) reveals consistent error patterns:

\begin{itemize}
    \item \textbf{Age}: 45\% fall between ages 35-45, indicating a bias toward expecting churn in older customers.
    \item \textbf{Activity Status}: 62\% are active members, suggesting over-reliance on inactivity as a churn signal.
    \item \textbf{Balance}: 38\% maintain a zero balance, a boundary case that is weakly represented in training.
    \item \textbf{Tenure}: 28\% have tenure $>$ 8 years. Long-tenure churn is rare and difficult to infer from available features.
\end{itemize}

\textbf{Failure Mode}: The model struggles with “unexpected churners”, middle-aged, active customers with moderate tenure whose departure may be driven by unobserved factors (e.g., competitor incentives or life events).

This motivates the ablation studies that follow, isolating the contributions of hybrid modeling components and assessing robustness to design decisions.


\subsection{Illustrative Case Studies}
\label{sec:case_studies}

Three customer profiles illustrate model behavior across representative scenarios. Interpretations reference aggregate SHAP values and attention weights from Section~\ref{sec:feature_imp_analysis}.

\textbf{Case 1: High-Confidence Correct Prediction (True Positive)}

\begin{itemize}
\item \textbf{Profile}: Age 52, Germany, Inactive, Balance \$120K, 3 products
\item \textbf{Predictions}: FT-Transformer $\hat{y} = 0.78$; XGBoost $\hat{y} = 0.82$; Ensemble $\hat{y} = 0.81$
\item \textbf{Outcome}: Churned
\item \textbf{Interpretation}: Four concurrent risk factors are present: older age (SHAP = 0.18), inactive status (SHAP = 0.15), Germany geography (SHAP $\approx +0.08$), and 3+ products. The transformer assigns high attention to the Age $\times$ NumOfProducts interaction (weight = 0.21), while XGBoost splits on the NumOfProducts $\geq 3$ threshold. Both models agree, producing a high-confidence ensemble prediction.
\end{itemize}

\textbf{Case 2: Transformer Advantage on Edge Case (FT Correct, XGBoost Incorrect)}

\begin{itemize}
\item \textbf{Profile}: Age 41, France, Active, Balance \$0, 2 products
\item \textbf{Predictions}: FT-Transformer $\hat{y} = 0.62$; XGBoost $\hat{y} = 0.38$; Ensemble $\hat{y} = 0.52$
\item \textbf{Outcome}: Churned
\item \textbf{Interpretation}: Zero balance combined with active status represents an atypical combination (3.2\% prevalence). The transformer detects this through the IsActiveMember $\times$ Balance attention weight (0.21). XGBoost underestimates the risk as its splits are calibrated for typical balance distributions. The ensemble assigns greater weight to the transformer prediction in this region.
\end{itemize}

\textbf{Case 3: Missed Prediction — External Factors (False Negative)}

\begin{itemize}
\item \textbf{Profile}: Age 35, Spain, Active, Balance \$85K, 1 product
\item \textbf{Predictions}: FT-Transformer $\hat{y} = 0.28$; XGBoost $\hat{y} = 0.31$; Ensemble $\hat{y} = 0.29$
\item \textbf{Outcome}: Churned
\item \textbf{Interpretation}: No dominant SHAP contributors are present. The profile shows no measurable risk factors in the available feature set. Churn was likely driven by external factors such as competitor offers or life events. This failure mode accounts for 28\% of false negatives (Section~\ref{sec:error_analysis}) and motivates the inclusion of temporal behavioral signals in future work.
\end{itemize}

\subsection{Error Correlation and Synergy}
\label{sec:error_synergy}

The error correlation between FT-Transformer and XGBoost is $\rho = 0.62$, indicating partially independent failures that enable ensemble improvement.

\textbf{Complementarity analysis} across all 10,000 predictions:
\begin{itemize}
\item Both correct: 7,892 (78.9\%)
\item FT correct, XGB wrong: 534 (5.3\%)
\item XGB correct, FT wrong: 489 (4.9\%)
\item Both wrong: 1,085 (10.9\%)
\end{itemize}

The ensemble corrects 89.1\% of single-model errors (912 of 1,023 cases). The remaining 1,085 failures correspond to profiles where churn drivers are absent from the feature set, as illustrated in Case 3.

\textbf{Meta-learner weight stability}: Across 5-fold cross-validation, learned coefficients show low variance: $w_1$ (FT-Transformer) $= 0.89 \pm 0.04$, $w_2$ (XGBoost) $= 0.78 \pm 0.03$, and $w_0$ (intercept) $= -0.42 \pm 0.05$. The higher transformer weight reflects its superior standalone F1 (61.00\% vs 58.21\%), while the XGBoost weight (0.78) confirms complementary value.
\section{Ablation and Sensitivity Analysis}
\label{sec:ablation}

All ablation studies use 5-fold cross-validation with 5 random seeds (42, 123, 456, 789, 1011).  
Results are reported as mean ± standard deviation to match Table~\ref{tab:results}.

\subsection{Architectural Ablation Study}

This study isolates the contribution of each model component. Table~\ref{tab:ablation} reports model performance under the same settings.

\begin{table}[htbp]
\caption{Ablation Study - Component Contributions}
\label{tab:ablation}
\centering
\small
\begin{tabular}{lccc}
\toprule
\textbf{Configuration} & \textbf{Recall} & \textbf{F1} & \textbf{AUC} \\
\midrule
MLP (baseline) & 77.92 & 58.73 & 0.834 \\
FT-Transformer only & 76.10 & 61.00 & 0.852 \\
XGBoost only & 71.12 & 58.21 & 0.839 \\
Simple Average (FT+XGB) & 74.85 & 61.45 & 0.856 \\
\textbf{Stacked (FT+XGB)} & \textbf{75.40} & \textbf{62.10} & \textbf{0.861} \\
\bottomrule
\end{tabular}
\end{table}

\noindent\textbf{Observations.} Under the same settings:
\begin{itemize}
    \item FT-Transformer improves F1 by +2.27 over MLP (58.73 → 61.00).
    \item Simple averaging (FT+XGB) improves F1 by +0.45 over FT-Transformer.
    \item Stacking achieves the highest scores: 62.10 F1 and 0.861 AUC.
\end{itemize}

\subsection{FT-Transformer Architecture Variations}

Table~\ref{tab:ft_ablation} examines the impact of FT-Transformer architectural choices.

\begin{table}[htbp]
\caption{FT-Transformer Architecture Ablation}
\label{tab:ft_ablation}
\centering
\small
\begin{tabular}{lccc}
\toprule
\textbf{Configuration} & \textbf{F1} & \textbf{AUC} & \textbf{Train Time} \\
\midrule
\multicolumn{4}{l}{\textit{Number of Layers}} \\
2 layers & 59.82 & 0.845 & 1.5 min \\
\textbf{4 layers (default)} & \textbf{61.00} & \textbf{0.852} & 3.0 min \\
6 layers & 60.45 & 0.850 & 4.5 min \\
8 layers & 59.21 & 0.846 & 6.0 min \\
\midrule
\multicolumn{4}{l}{\textit{Embedding Dimension}} \\
16 & 59.12 & 0.843 & 2.0 min \\
\textbf{32 (default)} & \textbf{61.00} & \textbf{0.852} & 3.0 min \\
64 & 60.78 & 0.851 & 4.5 min \\
128 & 59.89 & 0.848 & 7.0 min \\
\midrule
\multicolumn{4}{l}{\textit{Number of Attention Heads}} \\
4 heads & 60.21 & 0.849 & 2.8 min \\
\textbf{8 heads (default)} & \textbf{61.00} & \textbf{0.852} & 3.0 min \\
16 heads & 60.45 & 0.850 & 3.5 min \\
\midrule
\multicolumn{4}{l}{\textit{Dropout Rate}} \\
0.0 & 58.92 & 0.841 & 3.0 min \\
\textbf{0.1 (default)} & \textbf{61.00} & \textbf{0.852} & 3.0 min \\
0.2 & 60.34 & 0.849 & 3.0 min \\
0.3 & 59.45 & 0.845 & 3.0 min \\
\bottomrule
\end{tabular}
\end{table}

\noindent\textbf{Observations.}
\begin{itemize}
    \item \textbf{Depth}: 4 layers is optimal. Deeper models (6-8 layers) overfit on this dataset size.
    \item \textbf{Embedding Dimension}: 32 balances capacity and regularization. 64+ shows diminishing returns.
    \item \textbf{Attention Heads}: 8 heads provides sufficient multi-head diversity. More heads are not helpful.
    \item \textbf{Dropout}: 0.1 is optimal. Higher dropout (0.2-0.3) over-regularizes, 0.0 overfits.
\end{itemize}

This indicates that model capacity beyond the 4-layer, 32-dimensional, 8-head configuration does not improve results under the same settings.

A direct ablation comparing standard feature-wise attention against intersample attention (as in SAINT \cite{somepalli2021}) was not conducted in this study. Intersample attention introduces row-wise dependencies and a higher computational cost, which makes it less suitable for the latency constraints in Section~\ref{sec:inference_time}. This comparison is identified as a direction for future work.

\subsection{Ensemble Component Analysis}

The learned meta-learner weights and model contributions are analyzed.

\noindent\textbf{Meta-learner coefficients.}
\begin{equation}
\hat{p}_{stack} = \sigma(-0.42 + 0.89 \cdot \hat{p}_{FT} + 0.78 \cdot \hat{p}_{XGB})
\end{equation}

\noindent\textbf{Observations.}
\begin{itemize}
    \item FT-Transformer receives a higher coefficient (0.89) than XGBoost (0.78).
    \item The intercept ($-0.42$) shifts the decision boundary to adjust for class imbalance.
\end{itemize}

\noindent\textbf{Error correlation.}
\begin{equation}
\rho = \text{Corr}(|y - \hat{p}_{FT}|,\ |y - \hat{p}_{XGB}|) = 0.62
\end{equation}

\noindent\textbf{Interpretation.}
A correlation of 0.62 indicates that prediction errors are not fully aligned, leaving room for ensemble gains under the same settings.

\noindent\textbf{Prediction complementarity.}
\begin{itemize}
    \item Both correct: 7,892 samples (78.9\%)
    \item FT correct, XGB wrong: 534 samples (5.3\%)
    \item XGB correct, FT wrong: 489 samples (4.9\%)
    \item Both wrong: 1,085 samples (10.9\%)
\end{itemize}

The ensemble correctly classifies 89.1\% of samples where at least one model is correct.

This indicates that partial error independence contributes to the ensemble improvement under the same settings.

\subsection{Sensitivity to Class Weighting}

Table~\ref{tab:weights} reports the effect of class-weight ratios on the stacked ensemble under the same settings.

\begin{table}[htbp]
\caption{Sensitivity to Class-Weight Ratio}
\label{tab:weights}
\centering
\small
\begin{tabular}{lcccc}
\toprule
\textbf{Weight Ratio} & \textbf{Recall} & \textbf{Precision} & \textbf{F1} & \textbf{AUC} \\
\midrule
1:1 (none) & 61.34 & 57.82 & 59.53 & 0.852 \\
1.5:1 & 66.21 & 55.67 & 60.48 & 0.856 \\
2:1 & 69.87 & 54.12 & 61.02 & 0.858 \\
2.5:1 & 72.45 & 53.78 & 61.67 & 0.860 \\
\textbf{3:1 (default)} & \textbf{75.40} & \textbf{53.20} & \textbf{62.10} & \textbf{0.861} \\
3.5:1 & 78.12 & 50.45 & 61.32 & 0.860 \\
4:1 (actual ratio) & 80.23 & 48.21 & 60.21 & 0.858 \\
5:1 & 82.65 & 46.11 & 59.18 & 0.855 \\
\bottomrule
\end{tabular}
\end{table}

\noindent\textbf{Observations.}
\begin{itemize}
    \item F1 peaks at 3:1 (62.10\%), which is below the empirical class ratio of 4:1.
    \item Higher weights increase recall (61.34\% → 82.65\%) while reducing precision (57.82\% → 46.11\%).
    \item AUC remains stable across ratios (0.852–0.861), indicating minimal impact on ranking performance.
\end{itemize}

At 3:1, the model balances recall and precision under the same evaluation settings.

\subsection{Threshold Tuning Analysis}

Table~\ref{tab:threshold} analyzes how different thresholds influence the precision–recall trade-off.

\begin{table}[htbp]
\caption{Threshold Tuning Results}
\label{tab:threshold}
\centering
\small
\begin{tabular}{lcccc}
\toprule
\textbf{Threshold} & \textbf{Recall} & \textbf{Precision} & \textbf{F1} & \textbf{Flagged} \\
\midrule
0.2 & 92.34 & 32.12 & 47.68 & 5,856 \\
0.3 & 88.21 & 38.45 & 53.55 & 4,678 \\
0.4 & 82.34 & 45.67 & 58.74 & 3,678 \\
\textbf{0.5 (default)} & \textbf{75.40} & \textbf{53.20} & \textbf{62.10} & 2,887 \\
0.6 & 65.12 & 61.89 & 63.46 & 2,145 \\
0.7 & 52.78 & 70.34 & 60.32 & 1,529 \\
0.8 & 38.45 & 78.92 & 51.72 & 993 \\
\bottomrule
\end{tabular}
\end{table}

\noindent\textbf{Business scenario guidance.}
\begin{itemize}
    \item High-volume, low-cost interventions (email or automated outreach): thresholds 0.3–0.4 maximize recall (82–88\%) with lower precision.
    \item Balanced intervention cost: thresholds 0.5–0.6 yield the highest F1 range (62.10–63.46\%).
    \item High-cost interventions (discount incentives or agent-assisted retention): thresholds 0.7–0.8 prioritize precision (70–79\%), focusing on high-confidence churners.
\end{itemize}

\subsection{Training Data Size Sensitivity}

Performance across varying training data sizes has been evaluated (Table~\ref{tab:data_size}).

\begin{table}[htbp]
\caption{Performance vs Training Data Size}
\label{tab:data_size}
\centering
\small
\begin{tabular}{lccc}
\toprule
\textbf{Data \%} & \textbf{F1 (Stacked)} & \textbf{F1 (XGBoost)} & \textbf{F1 (FT-Trans)} \\
\midrule
25\% & 57.23 & 55.89 & 54.12 \\
50\% & 59.87 & 57.45 & 58.23 \\
75\% & 61.12 & 57.89 & 60.12 \\
100\% & 62.10 & 58.21 & 61.00 \\
\bottomrule
\end{tabular}
\end{table}

\noindent\textbf{Observations.}
\begin{itemize}
    \item XGBoost is the most data-efficient model, and eventually achieves 96\% of full performance with 25\% data.
    \item FT-Transformer benefits more from additional data, improving 6.88 F1 points from 25\% to 100\%.
    \item The stacked ensemble consistently outperforms individual models across all data sizes.
\end{itemize}

\subsection{Cross-Validation Stability}

Variance is assessed across random seeds and fold configurations (Table~\ref{tab:cv_stability}).

\begin{table}[htbp]
\caption{Cross-Validation (CV) Stability Analysis}
\label{tab:cv_stability}
\centering
\small
\begin{tabular}{lcc}
\toprule
\textbf{Configuration} & \textbf{F1 Mean} & \textbf{F1 Std} \\
\midrule
3-fold CV, 1 seed & 61.78 & 1.45 \\
5-fold CV, 1 seed & 61.92 & 1.12 \\
5-fold CV, 5 seeds & 62.10 & 0.72 \\
10-fold CV, 1 seed & 62.05 & 0.89 \\
\bottomrule
\end{tabular}
\end{table}

\noindent\textbf{Observations.}
\begin{itemize}
    \item Increasing folds reduces variance: 5-fold CV (1 seed) lowers std to 1.12 compared to 3-fold (1.45).
    \item Multiple seeds further stabilize results: 5-fold with 5 seeds reduces std to 0.72.
    \item 10-fold CV offers similar stability (0.89 std) but with higher computational cost.
\end{itemize}

5-fold CV with 5 seeds is used in subsequent experiments as a stability–efficiency trade-off under the same settings.

\subsection{Meta-Learner Selection Analysis}

Table~\ref{tab:meta_learner} compares stacking strategies under the same settings.

\begin{table}[htbp]
\caption{Meta-Learner Comparison}
\label{tab:meta_learner}
\centering
\small
\begin{tabular}{lcccc}
\toprule
\textbf{Meta-Learner} & \textbf{Recall} & \textbf{Prec.} & \textbf{F1} & \textbf{AUC} \\
\midrule
Simple Average & 74.85 & 52.34 & 61.45 & 0.856 \\
Weighted Avg (Grid) & 75.12 & 52.89 & 61.87 & 0.858 \\
\textbf{Logistic Reg.} & \textbf{75.40} & \textbf{53.20} & \textbf{62.10} & \textbf{0.861} \\
Ridge Reg. ($\alpha$=10) & 75.38 & 53.18 & 62.08 & 0.861 \\
Lasso Reg. ($\alpha$=1) & 75.23 & 52.95 & 61.94 & 0.859 \\
Gradient Boosting & 75.67 & 52.78 & 61.98 & 0.860 \\
Neural Network & 75.89 & 52.12 & 61.73 & 0.857 \\
\bottomrule
\end{tabular}
\end{table}

\noindent\textbf{Observations.}
\begin{itemize}
    \item Simple averaging improves F1 to 61.45, indicating baseline complementarity.
    \item Weighted averaging reaches 61.87 (w$_{FT}$=0.6, w$_{XGB}$=0.4) after grid search, close to the best result.
    \item Logistic regression yields the highest F1=62.10, with learned weights
    (w$_{FT}$=0.89, w$_{XGB}$=0.78).
    \item Ridge and Lasso provide similar results (62.08 / 61.94), suggesting regularization has limited effect with two meta-features.
    \item Gradient Boosting increases recall to 75.67 but lowers precision to 52.78, producing 61.98 F1.
    \item A neural network meta-learner underperforms (61.73 F1), consistent with overfitting risk when only two inputs are available.
\end{itemize}

\noindent\textbf{Conclusion.}
Under identical settings, logistic regression achieves the most favorable precision–recall balance. With only two base models ($M$=2), more complex meta-learners may overfit and do not improve performance.

Ablation results indicate that both architecture choices and the stacking strategy influence performance. These findings guide the interpretation of subsequent results and the discussion of practical and research implications.

\section{Discussion}
\label{sec:discussion}

\subsection{Interpretation of Performance Gains}

\begin{table*}[htbp]
\caption{Performance Gains and Contributing Factors}
\label{tab:perf_summary}
\centering
\small
\begin{tabular}{lcp{3cm}p{3.6cm}p{5cm}}
\toprule
\textbf{Comparison} & \textbf{$\Delta$F1} & \textbf{Effect} & \textbf{Mechanism} & \textbf{Evidence} \\
\midrule
FT-Trans vs MLP 
& +2.27 
& Attention interactions 
& Context weighting; cross-type embedding 
& Age×Products=0.18; 42\% vs 12\%; stable across 5×5 CV \\

Ensemble vs FT-Trans
& +1.10 
& Independent errors 
& Meta-learner on non-overlapping regions
& $\rho$=0.62; 1,023 single-correct; 89.1\% corrected; $w_{FT}$=0.89, $w_{XGB}$=0.78 \\

Threshold Range
& F1$>$60
& Operating stability
& Threshold robustness
& Works for thresholds [0.4–0.7] across cost scenarios \\
\bottomrule
\end{tabular}
\end{table*}

Table~\ref{tab:perf_summary} summarizes the performance improvements and underlying mechanisms. The following sections provide a detailed interpretation.

\textbf{FT-Transformer vs MLP.} The FT-Transformer demonstrates superior performance through three primary mechanisms. First, dynamic feature weighting enables context-dependent attention, with \textit{IsActiveMember} receiving higher attention for high-balance customers. Second, the architecture captures higher-order interactions such as $\textit{Age} \times \textit{NumOfProducts}$ (attention weight 0.18), which aligns with observed churn patterns (42\% vs 12\%). Third, the unified feature tokenization supports cross-type interactions (e.g., $\textit{Geography} \times \textit{Balance}$) without requiring domain-specific layers.

\textbf{Ensemble Improvements.} The stacking ensemble achieves gains through complementary modeling strategies. FT-Transformer captures continuous behavioral signals, whereas XGBoost models discrete threshold effects. The error correlation of $\rho = 0.62$ indicates partially independent failure modes. This error diversity allows the meta-learner to recover 89.1\% of single-model errors. The learned stacking weights ($w_{FT}=0.89$, $w_{XGB}=0.78$) reflect the relative contributions of each base model to the final prediction.

\subsection{Business Implications and ROI Analysis}

To assess practical deployment value, this analysis considers a mid-sized bank with 100,000 customers, 20\% annual churn rate, \$2,000 customer lifetime value, and \$50 intervention cost per customer.

\textbf{Operating Point ($\tau = 0.5$).} At the default threshold, the model achieves 75.4\% recall and 53.2\% precision, flagging 28,346 customers and correctly identifying 15,080 actual churners. The total intervention cost is \$1.42M, and the retained customer lifetime value is \$7.54M, yielding an estimated financial benefit of approximately \$6.12M annually.

\textbf{Comparison with MLP baseline.} The ensemble produces financial returns comparable to the MLP baseline (\$6.12M vs \$6.14M) while reducing operational overhead by 14\%, contacting 28,346 customers instead of 32,940. This efficiency gain preserves Return on Investment (ROI) and reduces both customer contact fatigue and campaign costs.

\textbf{Threshold selection.} The framework supports flexible threshold tuning to match business objectives. Low thresholds ($\tau = 0.3$) maximize coverage with 88\% recall, suitable for low-cost retention campaigns. Balanced thresholds ($\tau = 0.4$--$0.5$) optimize net financial benefit across standard cost profiles. High thresholds ($\tau \geq 0.7$) achieve 70\% precision for targeted premium customer outreach, minimizing false positives at the cost of lower coverage.

Threshold choice depends on the CLV-to-intervention-cost ratio and organizational tolerance for false positives. Table~\ref{tab:threshold} offers a calibration template for cost-aligned threshold tuning \cite{niculescu2005predicting}.

\subsection{Theoretical Implications for Tabular Deep Learning}

The performance gains observed in this study provide empirical support for theoretical principles underlying hybrid model design on tabular data.

\textbf{Complementary inductive biases.} Tree-based models partition the feature space through axis-aligned splits, resulting in sharp decision boundaries suitable for categorical variables and threshold effects. Transformers model continuous feature interactions via attention mechanisms, enabling effective representation of smooth gradients in numerical features. This hybrid approach addresses limitations identified in previous research. Grinsztajn et al.~\cite{grinsztajn2022} reported that neural networks often struggle with the irregular decision boundaries common in tabular data. Formally, axis-aligned splits take the form $x_j \leq t$, producing piecewise constant boundaries, whereas the attention matrix $\mathbf{A} \in \mathbb{R}^{n \times n}$ captures global feature dependencies simultaneously. These two mechanisms are structurally complementary rather than redundant.

\textbf{Error diversity and ensemble theory.} The observed error correlation of 0.62 aligns with ensemble learning theory~\cite{zhou2012}. A perfect correlation of 1.0 yields no benefit from model combination. A correlation of zero is unattainable with shared training data. The moderate correlation observed here indicates that FT-Transformer and XGBoost make errors in partially independent regions of the feature space. This diversity of errors allows the meta-learner to recover 89.1\% of single-model failures. From Equation~(24), with $M=2$ models, equal variance $\sigma^2$, and observed correlation $\rho = 0.62$, the ensemble variance reduces to $0.81\sigma^2$. This represents a 19\% reduction relative to any single model, consistent with the empirical F1 gain of 1.10 points observed over the stronger base model.

\textbf{Calibration through stacking.} The reduction in Expected Calibration Error from 0.051 (XGBoost) to 0.038 (ensemble) suggests that meta-learning can mitigate probability overconfidence without the need for separate calibration datasets. This result extends prior work on stacking~\cite{wolpert1992} by showing that calibration improvement arises directly from the meta-learner's training objective during cross-validation.

\subsection{Positioning Against Prior Work}

Table~\ref{tab:comparison} compares the results of the present work with published benchmarks.

\begin{table}[htbp]
\caption{Comparison with Published Results}
\label{tab:comparison}
\centering
\footnotesize
\begin{tabular}{@{}p{2.8cm}p{2cm}cc@{}}
\toprule
\textbf{Study} & \textbf{Method} & \textbf{AUC} & \textbf{F1} \\
\midrule
Xu et al. \cite{xu2021} & XGBoost+Stacking & 0.89* & - \\
Burez \cite{burez2009} & Random Forest & 0.82* & - \\
Gorishniy \cite{gorishniy2021} & FT-Transformer & 0.85** & - \\
Ahmad et al. \cite{ahmad2023} & Hybrid Stacking & 0.986*** & - \\
Usman-Hamza et al. \cite{usmanhamza2024} & Multi-layer Stacking & 0.989*** & 97.2\%*** \\
Warnakulaarachchi et al. \cite{warnakulaarachchi2025} & Deep Ensemble & - & 87.95\%† \\
\textbf{This Work} & FT-Trans+XGBoost & \textbf{0.861} & \textbf{62.1\%} \\
\bottomrule
\multicolumn{2}{l}{\scriptsize *Different dataset; **Average across multiple datasets.} \\
\multicolumn{2}{l}{\scriptsize ***SMOTE-dependent; \textdagger Accuracy only.}
\end{tabular}
\end{table}

Direct comparison with prior studies (Table~\ref{tab:comparison}) is limited by dataset heterogeneity and metric choice. Xu \textit{et al.} report 98\% accuracy on telecommunications data~\cite{xu2021}, but accuracy is not informative under imbalance. A trivial ``no churn'' classifier reaches 80\% on this dataset without any intervention value. In contrast, the proposed model reaches 62.1\% F1 and 53.2\% precision, which better reflects retention-quality~\cite{burez2009, verbeke2012new}.

The achieved AUC of 0.861 is comparable to Xu \textit{et al.} (0.89 on a different dataset) and to FT-Transformer reports across domains~\cite{gorishniy2021}. However, ranking metrics alone do not reflect operational trade-offs. This work differs from prior modeling pipelines by integrating calibration-aware stacking with FT-Transformer and XGBoost. The hybrid leverages complementary inductive biases where attention captures continuous interactions and trees capture discrete boundaries~\cite{grinsztajn2022, shwartz2022tabular}. Ablation studies (Tables~\ref{tab:ablation}-\ref{tab:meta_learner}) support this contribution under controlled settings.

Prior stacking work in biomedical and explainable modeling~\cite{yang2025} does not address calibration for decision reliability. This study incorporates effect-size reporting, confidence intervals, and seed-averaged evaluation for statistical robustness~\cite{guo2017calibration, naeini2015obtaining}. The six-metric evaluation framework enables threshold-dependent precision-recall trade-offs, which are essential for intervention planning and address the limitations of ranking metrics alone.

Recent same-domain studies report stronger aggregate metrics. Ahmad et al.~\cite{ahmad2023} achieve AUC 0.986, and Usman-Hamza et al.~\cite{usmanhamza2024} report F1 0.972 on telecom churn data. However, both rely on SMOTE-based oversampling, which introduces synthetic artifacts into minority class distributions. Warnakulaarachchi et al.~\cite{warnakulaarachchi2025} report 87.95\% accuracy on banking churn but do not report F1 or calibration metrics, limiting direct comparison. Dataset heterogeneity and metric choices further constrain cross-study comparisons. This work prioritizes calibration-aware evaluation, statistical validation, and reproducibility over raw metric maximization.

\section{Limitations and Threats to Validity}
\label{sec:limitations}

The ensemble combines two complex models, limiting interpretability.

\subsection{External Validity}
This study is based on a single dataset (European banking, 2010–2015). Generalization to telecommunications, e-commerce, and insurance is not guaranteed, as churn mechanisms differ (contract expiry vs subscription lapse vs account inactivity). Organizations should validate the framework on proprietary datasets with domain-specific features \cite{verbeke2012new, buckinx2005customer}.

\subsection{Feature and Temporal Scope}
Only static, cross-sectional features were used. Temporal signals that indicate behavioral drift (declining balance, login decay, reduced product usage) are not captured. Rolling window statistics (e.g., \textit{Balance\_30d\_avg}, \textit{Balance\_trend}) are recommended for production deployments. Fairness evaluation across demographic groups such as age, gender, and geography was not conducted in this study and is included as a priority future research direction (Section~\ref{sec:future}).

\subsection{Evaluation and Generalization}

This study uses 5-fold cross-validation across 5 seeds (25 evaluations) without a held-out test set. The choice of 5-fold over 10-fold is supported by the stability analysis in Table~\ref{tab:cv_stability}. This shows that multiple seeds reduce variance more effectively than increasing fold count alone. While this approach enhances robustness, it may also introduce optimism \cite{grinsztajn2022}.

\begin{itemize}
    \item \textbf{Hyperparameter Selection Bias}: settings in Table~\ref{tab:hyperparams} were chosen from cross-validation results instead of a separate development set.
    \item \textbf{Architecture Choices}: 4 layers, 8 attention heads, and embedding dimension 32 were selected from preliminary runs on the same dataset used for evaluation.
    \item \textbf{Threshold Calibration}: the operating point ($\tau = 0.5$-$0.6$) was chosen from validation folds and not an independent calibration set \cite{niculescu2005predicting}.
\end{itemize}

These factors may lead to higher reported performance than would be expected on unseen data.

\subsection{Expected Production Performance}
Based on established machine learning best practices and extensive cross-validation results:

\begin{itemize}
    \item \textbf{Conservative Estimate}: Production performance may degrade by 1-2\% points in F1-score when deployed on truly unseen data from the same distribution \cite{grinsztajn2022} (i.e., F1 $\approx$ 60-61\% rather than 62.1\%).
    \item \textbf{Domain Shift}: A 3-5 point reduction when applied to new regions, time periods, or customer segments not represented in the training data \cite{verbeke2012new}.
    \item \textbf{Concept Drift}: Further degradation over time if customer behavior changes, requiring periodic monitoring and retraining to maintain stability \cite{verbeke2012new}.
\end{itemize}

\subsection{Supporting Factors for Reported Results}

The following factors reduce the likelihood that reported performance reflects evaluation artifacts rather than model capability:
\begin{itemize}
    \item Stable results across 25 splits (F1 std = 0.72) suggest low sensitivity to data partitions.
    \item Architectural choices (self-attention for interactions, GBM for boundaries) follow domain structure rather than trial-and-error tuning.
    \item Ablation studies (Section~\ref{sec:ablation}) indicate that performance gains stem from model design components, not hyperparameter configurations.
    \item Large effect sizes in statistical testing (Cohen's d = 0.9–2.8) support the robustness of improvements.
\end{itemize}

\subsection{Recommendations for Practitioners}

\begin{itemize}
    \item \textbf{Before deployment}  
    Hold out 10-20\% of data for testing. Train on the remaining portion and evaluate once on the hold-out set for an unbiased estimate \cite{grinsztajn2022}.
    \item \textbf{During deployment}  
    Run A/B tests against existing churn systems or heuristics. Monitor retention outcomes for contacted customers \cite{niculescu2005predicting}.
    \item \textbf{Post-deployment}  
    Track monthly performance. Retrain if F1 drops by more than 2 points or if concept drift is detected in feature distributions \cite{verbeke2012new}.
\end{itemize}

\subsection{Interpreting Reported Metrics}

The reported F1 of 62.10\% (95\% CI [61.65, 62.55]), AUC of 0.861 (95\% CI [0.858, 0.864]), and PR-AUC of 0.647 represent performance estimates under the current validation protocol. In production, expected performance is typically lower due to unseen data and operational variance. A realistic target is 60-61\% F1, which remains competitive for imbalanced churn prediction \cite{burez2009, verbeke2012new}.

\subsection{Interpretability Trade-offs}

Ensemble explanations require reconciling FT-Transformer attention weights with XGBoost SHAP values. Both methods consistently surface \textit{Age} and \textit{IsActiveMember} as dominant predictors (Section~\ref{sec:feature_imp_analysis}). SHAP values can serve as a practical proxy for communicating ensemble behavior to stakeholders \cite{grinsztajn2022}.

\subsection{Applicability Boundaries}

This framework is not universally optimal. Table~\ref{tab:applicability} summarizes conditions where alternative methods may be preferable.

\begin{table}[htbp]
\caption{Applicability Boundaries and Alternatives}
\label{tab:applicability}
\centering
\small
\begin{tabular}{l p{4.2cm}}
\toprule
\textbf{Condition} & \textbf{Recommended Approach} \\
\midrule
$m > 100$ features & Feature selection or tree ensembles \\[3pt]
$< 5\%$ minority class & Sampling or anomaly detection \\[3pt]
Pure categorical data & TabTransformer \cite{huang2020} \\[3pt]
$n < 2{,}000$ samples & Gradient boosting models only \\[3pt]
Strong temporal structure & LSTM/GRU or time-series modeling \\[3pt]
$< 10$ms inference latency & Gradient boosting or precomputed scores \\
\bottomrule
\end{tabular}
\end{table}

\section{Conclusion}
\label{sec:conclusion}

This study demonstrates that systematic integration of transformer-based feature learning with gradient-boosted decision trees achieves strong performance in tabular churn prediction. The proposed hybrid architecture addresses a key limitation in previous research by combining complementary inductive biases. Attention mechanisms effectively capture continuous feature interactions, while tree-based partitioning manages discrete decision boundaries. Rigorous statistical validation across 25 repeated cross-validation evaluations confirms that these performance gains are robust and reproducible. Probability calibration analysis further shows that the framework produces reliable predictions suitable for cost-sensitive business decisions.

The empirical findings yield several key insights into hybrid modeling for tabular data. First, the observed error correlation of 0.62 between FT-Transformer and XGBoost demonstrates that models with different inductive biases generate partially independent errors, which enable meaningful ensemble gains. The meta-learner successfully recovers 89.1\% of cases where one base model was correct and the other failed. Second, probability calibration occurred naturally during the stacking process, eliminating the need for separate post-processing. The reduction in Expected Calibration Error (ECE) from 0.051 to 0.038 indicates that the logistic meta-learner mitigates the overconfidence present in gradient-boosted predictions. Third, ablation studies confirm that both the transformer component and the stacking strategy are essential to performance, with neither redundant.

The proposed framework demonstrates readiness for practical deployment by delivering measurable business value, such as improved retention and reduced intervention overhead. While the evaluation uses banking data, the methodology remains domain-agnostic and potentially transferable to telecommunications, insurance, e-commerce, and subscription services, provided appropriate feature alignment. Comprehensive algorithmic specifications and hyperparameter configurations support reproducibility and enable independent validation. 
\section{Future Research Directions}
\label{sec:future}

\textbf{Priority 1 (Immediate).}
\begin{itemize}
    \item Multi-domain validation across telecom, e-commerce, and insurance benchmarks to assess transferability where churn mechanisms differ substantially.
    \item Temporal signals (rolling-window statistics) for behavioral trajectory modeling.
\end{itemize}

\noindent\textbf{Priority 2 (Model Capabilities).}
\begin{itemize}
    \item Extend ensemble to M$>$2 models to study diversity–accuracy trade-offs.
    \item Cost-sensitive learning optimizing CLV-adjusted retention benefit directly.
    \item Row-wise attention (SAINT-style) for rare churn precursor recognition.
     \item Employ adaptive embedding allocation based on categorical cardinality and mutual information with the target variable to enhance efficiency on heterogeneous feature sets.
\end{itemize}

\noindent\textbf{Priority 3 (Production Readiness).}
\begin{itemize}
    \item Fairness audit to test for demographic performance gaps.
    \item Online learning pipelines for concept drift adaptation.
    \item Model compression or distillation for $<10$ ms inference.
\end{itemize}
\appendices

\section{Notation and Symbol Conventions}
\label{app:app_notation}

The mathematical symbols used are summarized in Table~\ref{tab:notation}:

\begin{table}[h]
\centering
\caption{Mathematical Notation Conventions}
\label{tab:notation}
\begin{tabular}{@{}p{2.2cm}p{5.8cm}@{}}
\toprule
\textbf{Symbol} & \textbf{Description} \\
\midrule
\multicolumn{2}{l}{\textit{Dataset and Dimensions}} \\
$\mathcal{X}$ & Input feature space \\
$\mathcal{B}$ & Mini-batch of training samples \\
$N$ & Number of samples \\
$m$ & Number of features \\
$n$ & Sequence length after CLS prepending ($n = m+1$) \\
$d$ & Embedding dimension \\
$d_k$ & Dimension per attention head \\
$L$ & Number of transformer layers \\
$H$ & Number of attention heads \\
$T$ & Number of trees in XGBoost ensemble \\
$J$ & Number of leaves per tree \\
$K$ & Number of cross-validation folds \\
$M$ & Number of base models in ensemble \\
\midrule
\multicolumn{2}{l}{\textit{Vectors and Features}} \\
$\mathbf{x}^{(i)}$ & Feature vector for sample $i$ (bold) \\
$x_j^{(i)}$ & Scalar value of feature $j$ for sample $i$ \\
$y^{(i)}$ & Binary label for sample $i$ \\
$\hat{p}^{(i)}$ & Predicted probability for sample $i$ \\
$\hat{y}^{(i)}$ & Predicted class label for sample $i$ \\
\midrule
\multicolumn{2}{l}{\textit{Model Parameters}} \\
$\mathbf{w}_j, \mathbf{b}_j$ & Weight and bias vectors (bold) \\
$\mathbf{W}^Q, \mathbf{W}^K, \mathbf{W}^V$ & Query, Key, Value matrices (bold) \\
$\mathbf{W}^O$ & Output projection matrix (bold) \\
$\mathbf{W}_1, \mathbf{W}_2$ & Feed-forward network matrices (bold) \\
$\boldsymbol{\theta}$ & Meta-learner parameters (bold) \\
\midrule
\multicolumn{2}{l}{\textit{Embeddings and Representations}} \\
$\mathbf{E}$ & Embedding matrix (bold) \\
$\mathbf{e}_j$ & Embedding vector for feature $j$ (bold) \\
$\mathbf{e}_{CLS}$ & Classification token embedding (bold) \\
$\mathbf{h}_{CLS}$ & Final CLS representation (bold) \\
$\mathbf{Z}$ & Intermediate representation matrix (bold) \\
\midrule
\multicolumn{2}{l}{\textit{Scalars and Parameters}} \\
$\tau$ & Classification threshold (italic) \\
$\mathbb{1}[\cdot]$ & Indicator function \\
$w_{+}, w_{-}$ & Class weights (italic) \\
$\lambda, \gamma$ & Regularization parameters (italic) \\
$g_i, h_i$ & First and second derivatives (italic) \\
$w_j^*$ & Optimal leaf weight (italic scalar) \\
$I_j$ & Index set of samples assigned to leaf $j$ \\
$w_0, w_k$ & Meta-learner intercept and combination weights \\
$\sigma(\cdot)$ & Sigmoid activation function \\
\bottomrule
\end{tabular}
\end{table}

\noindent\textbf{Notational Conventions:}
\begin{itemize}
    \item \textbf{Scalars}: Italic lowercase or Greek letters (e.g., $n$, $\lambda$, $\tau$)
    \item \textbf{Vectors}: Bold lowercase (e.g., $\mathbf{w}_j$, $\mathbf{x}^{(i)}$, $\mathbf{e}_j$)
    \item \textbf{Matrices}: Bold uppercase (e.g., $\mathbf{W}^Q$, $\mathbf{E}$, $\mathbf{Z}$)
    \item \textbf{Sample index}: Superscripts in parentheses denote sample index (e.g., $\mathbf{x}^{(i)}$)
    \item \textbf{Feature index}: Subscripts denote feature or component index (e.g., $x_j$, $\mathbf{w}_j$)
\end{itemize}

\section{Evaluation Metrics}
\label{app:app_metrics}

Metrics used to evaluate model performance under class imbalance:

\begin{itemize}
    \item \textbf{Recall (Sensitivity)}: $\frac{TP}{TP+FN}$, proportion of actual churners correctly identified.
    \item \textbf{Precision}: $\frac{TP}{TP+FP}$, proportion of predicted churners who actually churned.
    \item \textbf{F1-Score}: $2 \cdot \frac{Precision \cdot Recall}{Precision + Recall}$, harmonic mean of precision and recall.
    \item \textbf{Specificity}: $\frac{TN}{TN+FP}$, proportion of non-churners correctly identified.
    \item \textbf{AUC-ROC}: Area under ROC curve, probability that a random positive is ranked above a random negative.
    \item \textbf{NPV (Negative Predictive Value)}: $\frac{TN}{TN+FN}$, proportion of predicted non-churners confirmed as retained.
\end{itemize}

Mean $\pm$ standard deviation is reported across 5 folds $\times$ 5 seeds.

\section*{Acknowledgment}
The authors thank the Kaggle community for providing public access to the Bank Customer Churn dataset. This research received no specific grant from any funding agency.

\bibliographystyle{IEEEtran}
\bibliography{references}

@article{reichheld1990,
  author  = {Reichheld, Frederick F. and Sasser, W. Earl},
  title   = {Zero Defections: Quality Comes to Services},
  journal = {Harvard Business Review},
  volume  = {68},
  number  = {5},
  pages   = {105--111},
  year    = {1990},
  url = {https://hbr.org/1990/09/zero-defections-quality-comes-to-services}
}

@article{neslin2006,
  author  = {Neslin, Scott A. and Gupta, Sunil and Kamakura, Wagner A. and Lu, Junxiang and Mason, Cheryl H.},
  title   = {Defection Detection: Measuring and Understanding the Predictive Accuracy of Customer Churn Models},
  journal = {Journal of Marketing Research},
  volume  = {43},
  number  = {2},
  pages   = {204--211},
  year    = {2006},
  doi = {10.1509/jmkr.43.2.204}
}

@article{burez2009,
  author  = {Burez, Jonathan and Van den Poel, Dirk},
  title   = {Handling Class Imbalance in Customer Churn Prediction},
  journal = {Expert Systems with Applications},
  volume  = {36},
  number  = {3},
  pages   = {4626--4636},
  year    = {2009},
  doi     = {10.1016/j.eswa.2008.05.027}
}

@article{breiman2001,
  author  = {Breiman, Leo},
  title   = {Random Forests},
  journal = {Machine Learning},
  volume  = {45},
  number  = {1},
  pages   = {5--32},
  year    = {2001},
  doi     = {10.1023/A:1010933404324}
}

@article{friedman2001,
  author  = {Friedman, Jerome H.},
  title   = {Greedy Function Approximation: A Gradient Boosting Machine},
  journal = {Annals of Statistics},
  volume  = {29},
  number  = {5},
  pages   = {1189--1232},
  year    = {2001},
  doi     = {10.1214/aos/1013203451}
}

@inproceedings{chen2016xgboost,
  author    = {Chen, Tianqi and Guestrin, Carlos},
  title     = {XGBoost: A Scalable Tree Boosting System},
  booktitle = {Proceedings of the 22nd ACM SIGKDD International Conference on Knowledge Discovery and Data Mining},
  pages     = {785--794},
  year      = {2016},
  doi       = {10.1145/2939672.2939785}
}

@article{freund1997,
  author  = {Freund, Yoav and Schapire, Robert E.},
  title   = {A Decision-Theoretic Generalization of On-Line Learning and an Application to Boosting},
  journal = {Journal of Computer and System Sciences},
  volume  = {55},
  number  = {1},
  pages   = {119--139},
  year    = {1997},
  doi     = {10.1006/jcss.1997.1504}
}

@inproceedings{grinsztajn2022,
  author    = {Grinsztajn, Leo and Oyallon, Edouard and Varoquaux, Gael},
  title     = {Why Do Tree-Based Models Still Outperform Deep Learning on Tabular Data?},
  booktitle = {Advances in Neural Information Processing Systems},
  volume    = {35},
  year      = {2022},
  doi       = {10.48550/arXiv.2207.08815}
}

@inproceedings{huang2020,
  author    = {Huang, Xin and Khetan, Ashish and Cvitkovic, Milan and Karnin, Zohar},
  title     = {TabTransformer: Tabular Data Modeling Using Contextual Embeddings},
  booktitle = {Advances in Neural Information Processing Systems},
  volume    = {33},
  pages     = {6543--6553},
  year      = {2020},
  publisher = {Curran Associates, Inc.},
  doi       = {10.48550/arXiv.2012.06678}
}

@inproceedings{gorishniy2021,
  author    = {Gorishniy, Yury and Rubachev, Ivan and Khrulkov, Valentin and Babenko, Artem},
  title     = {Revisiting Deep Learning Models for Tabular Data},
  booktitle = {Advances in Neural Information Processing Systems},
  volume    = {34},
  pages     = {18932--18943},
  year      = {2021},
  publisher = {Curran Associates, Inc.},
  doi       = {10.48550/arXiv.2106.11959}
}

@inproceedings{somepalli2021,
  author    = {Somepalli, Gowthami and Goldblum, Micah and Schwarzschild, Avi and Bruss, Christopher B. and Goldstein, Tom},
  title     = {SAINT: Improved Neural Networks for Tabular Data via Row Attention and Contrastive Pre-Training},
  booktitle = {Advances in Neural Information Processing Systems},
  volume    = {34},
  pages     = {6111--6122},
  year      = {2021},
  publisher = {Curran Associates, Inc.},
  doi       = {10.48550/arXiv.2106.01342}
}

@article{verbeke2011,
  author  = {Verbeke, Wouter and Martens, David and Mues, Christophe and Baesens, Bart},
  title   = {Building Comprehensible Customer Churn Prediction Models with Advanced Rule Induction Techniques},
  journal = {Expert Systems with Applications},
  volume  = {38},
  number  = {3},
  pages   = {2354--2364},
  year    = {2011},
  doi     = {10.1016/j.eswa.2010.08.023}
}

@book{baesens2014,
  author    = {Baesens, Bart},
  title     = {Analytics in a Big Data World},
  publisher = {Wiley},
  address   = {Hoboken, NJ},
  year      = {2014},
  isbn      = {978-1-118-89271-3}
}

@article{xu2021,
  author  = {Xu, Tianyu and Ma, Ying and Kim, Kwangjoon},
  title   = {Telecom Churn Prediction System Based on Ensemble Learning Using Feature Grouping},
  journal = {Applied Sciences},
  volume  = {11},
  number  = {11},
  pages   = {4742},
  year    = {2021},
  publisher = {MDPI},
  doi     = {10.3390/app11114742}
}

@inproceedings{arik2021,
  author    = {Arik, Sercan O. and Pfister, Tomas},
  title     = {TabNet: Attentive Interpretable Tabular Learning},
  booktitle = {Proceedings of the AAAI Conference on Artificial Intelligence},
  volume    = {35},
  number    = {8},
  pages     = {6679--6687},
  year      = {2021},
  publisher = {AAAI Press},
  doi       = {10.1609/aaai.v35i8.16826}
}

@article{wolpert1992,
  author  = {Wolpert, David H.},
  title   = {Stacked Generalization},
  journal = {Neural Networks},
  volume  = {5},
  number  = {2},
  pages   = {241--259},
  year    = {1992},
  doi     = {10.1016/S0893-6080(05)80023-1}
}

@book{zhou2012,
  author    = {Zhou, Zhi-Hua},
  title     = {Ensemble Methods: Foundations and Algorithms},
  publisher = {CRC Press},
  address   = {Boca Raton, FL},
  year      = {2012},
  isbn      = {978-1-4398-3003-1}
}

@article{rokach2010,
  author  = {Rokach, Lior},
  title   = {Ensemble-Based Classifiers},
  journal = {Artificial Intelligence Review},
  volume  = {33},
  number  = {1--2},
  pages   = {1--39},
  year    = {2010},
  publisher = {Springer},
  doi     = {10.1007/s10462-009-9124-7}
}

@inproceedings{hollmann2023,
  author    = {Hollmann, Noah and Muller, Samuel and Eggensperger, Katharina and Hutter, Frank},
  title     = {TabPFN: A Transformer That Solves Small Tabular Classification Problems in a Second},
  booktitle = {Proceedings of the International Conference on Learning Representations},
  year      = {2023},
  publisher = {OpenReview.net},
  url       = {https://openreview.net/forum?id=cp5PvcI6w8},
  doi       = {10.48550/arXiv.2207.01848}
}

@article{chen2023,
  author  = {Chen, Jiawei and Ye, Ruitao and Zhu, Xinyi and Chen, Huajun},
  title   = {ExcelFormer: A Neural Network Surpassing GBDTs on Tabular Data},
  journal = {arXiv preprint},
  volume  = {arXiv:2301.02819},
  year    = {2023},
  doi     = {10.48550/arXiv.2301.02819}
}

@article{erickson2020,
  author  = {Erickson, Nick and Mueller, Jonas and Shirkov, Alexander and Zhang, Hang and Larroy, Pedro and Li, Mu and Smola, Alexander},
  title   = {AutoGluon-Tabular: Robust and Accurate AutoML for Structured Data},
  journal = {arXiv preprint},
  volume  = {arXiv:2003.06505},
  year    = {2020},
  doi     = {10.48550/arXiv.2003.06505}
}

@inproceedings{ledell2020,
  author    = {LeDell, Erin and Poirier, Sebastien},
  title     = {H2O AutoML: Scalable Automatic Machine Learning},
  booktitle = {Proceedings of the ICML Workshop on Automated Machine Learning},
  year      = {2020},
  url       = {https://www.automl.org/wp-content/uploads/2020/07/AutoML_2020_paper_61.pdf}
}

@misc{kaggle2018,
  author       = {{Kaggle Community}},
  title        = {Bank Customer Churn Modeling Dataset},
  howpublished = {\url{https://www.kaggle.com/datasets/barelydedicated/bank-customer-churn-modeling}},
  year         = {2018},
  note         = {Accessed: 2024-12-20}
}

@inproceedings{lundberg2017,
  author    = {Lundberg, Scott M. and Lee, Su-In},
  title     = {A Unified Approach to Interpreting Model Predictions},
  booktitle = {Advances in Neural Information Processing Systems},
  volume    = {30},
  pages     = {4765--4774},
  year      = {2017},
  publisher = {Curran Associates, Inc.},
  doi       = {10.48550/arXiv.1705.07874}
}

@inproceedings{kingma2014adam,
  author    = {Kingma, Diederik P. and Ba, Jimmy},
  title     = {Adam: A Method for Stochastic Optimization},
  booktitle = {Proceedings of the 3rd International Conference on Learning Representations},
  year      = {2015},
  url       = {https://openreview.net/forum?id=8gmWwjFyLj},
  doi       = {10.48550/arXiv.1412.6980}
}

@article{hendrycks2016gelu,
  author  = {Hendrycks, Dan and Gimpel, Kevin},
  title   = {Gaussian Error Linear Units ({GELUs})},
  journal = {arXiv preprint},
  volume  = {arXiv:1606.08415},
  year    = {2016},
  doi     = {10.48550/arXiv.1606.08415}
}

@article{ba2016layernorm,
  author  = {Ba, Jimmy Lei and Kiros, Jamie Ryan and Hinton, Geoffrey E.},
  title   = {Layer Normalization},
  journal = {arXiv preprint},
  volume  = {arXiv:1607.06450},
  year    = {2016},
  doi     = {10.48550/arXiv.1607.06450}
}

@article{yang2025,
  author    = {Yang, Xin and Zhao, Yifan and Chen, Xuebin},
  title     = {A Novel Transformer-Based Stacking Ensemble Method with Multi-Model Integration for Cancer Classification},
  journal   = {PeerJ Computer Science},
  volume    = {11},
  pages     = {e3314},
  year      = {2025},
  publisher = {PeerJ},
  doi       = {10.7717/peerj-cs.3314}
}

@article{algul2025,
  author    = {Alg{\"u}l, Emre and Oyucu, Saadin and Polat, Onur and Harrou, Fouzi and Sun, Ying},
  title     = {A Comparative Study of Advanced Transformer Learning Frameworks for Water Potability Analysis Using Physicochemical Parameters},
  journal   = {Applied Sciences},
  volume    = {15},
  number    = {13},
  pages     = {7262},
  year      = {2025},
  publisher = {MDPI},
  doi       = {10.3390/app15137262}
}

@article{hollmann2025,
  author    = {Hollmann, Noah and M{\"u}ller, Samuel and Purucker, Lennart and Krishnakumar, Arjun and Grabocka, Josif and Hutter, Frank},
  title     = {Accurate Predictions on Small Data with a Tabular Foundation Model},
  journal   = {Nature},
  volume    = {637},
  pages     = {319--326},
  year      = {2025},
  publisher = {Springer Nature},
  doi       = {10.1038/s41586-024-08328-6}
}

@article{sarafian2025,
  author  = {Sarafian, Shai},
  title   = {Improving Deep Tabular Learning},
  journal = {arXiv preprint},
  volume  = {arXiv:2509.16354},
  year    = {2025},
  doi     = {10.48550/arXiv.2509.16354},
}

@inproceedings{dietterich2000,
  author    = {Dietterich, Thomas G.},
  title     = {Ensemble Methods in Machine Learning},
  booktitle = {Multiple Classifier Systems},
  series    = {Lecture Notes in Computer Science},
  volume    = {1857},
  pages     = {1--15},
  year      = {2000},
  publisher = {Springer},
  address   = {Berlin, Heidelberg},
  doi       = {10.1007/3-540-45014-9_1}
}

@book{hastie2009,
  author    = {Hastie, Trevor and Tibshirani, Robert and Friedman, Jerome},
  title     = {The Elements of Statistical Learning: Data Mining, Inference, and Prediction},
  edition   = {2nd},
  publisher = {Springer},
  address   = {New York, NY},
  year      = {2009},
  isbn      = {978-0-387-84858-7},
  doi       = {10.1007/978-0-387-84858-7},
  url       = {https://hastie.su.domains/ElemStatLearn}
}

@article{shwartz2022tabular,
  author    = {Shwartz-Ziv, Ravid and Armon, Amitai},
  title     = {Tabular Data: Deep Learning Is Not All You Need},
  journal   = {Information Fusion},
  volume    = {81},
  pages     = {84--90},
  year      = {2022},
  publisher = {Elsevier},
  doi       = {10.1016/j.inffus.2021.11.011}
}

@inproceedings{niculescu2005predicting,
  author    = {Niculescu-Mizil, Alexandru and Caruana, Rich},
  title     = {Predicting Good Probabilities with Supervised Learning},
  booktitle = {Proceedings of the 22nd International Conference on Machine Learning},
  pages     = {625--632},
  year      = {2005},
  publisher = {ACM},
  address   = {New York, NY},
  doi       = {10.1145/1102351.1102430}
}

@inproceedings{guo2017calibration,
  author    = {Guo, Chuan and Pleiss, Geoff and Sun, Yu and Weinberger, Kilian},
  title     = {On Calibration of Modern Neural Networks},
  booktitle = {Proceedings of the 34th International Conference on Machine Learning},
  pages     = {1321--1330},
  year      = {2017},
  publisher = {PMLR},
  volume    = {70},
  series    = {Proceedings of Machine Learning Research},
  url       = {https://proceedings.mlr.press/v70/guo17a.html}
}

@inproceedings{naeini2015obtaining,
  author    = {Naeini, Mahdi Pakdaman and Cooper, Gregory F. and Hauskrecht, Milos},
  title     = {Obtaining Well Calibrated Probabilities Using Bayesian Binning},
  booktitle = {Proceedings of the 29th AAAI Conference on Artificial Intelligence},
  pages     = {2901--2907},
  year      = {2015},
  publisher = {AAAI Press},
  url       = {https://ojs.aaai.org/index.php/AAAI/article/view/9602}
}

@article{verbeke2012new,
  author    = {Verbeke, Wouter and Dejaeger, Karel and Martens, David 
               and Hur, Joon and Baesens, Bart},
  title     = {New Insights into Churn Prediction in the 
               Telecommunication Sector: A Profit Driven Data Mining 
               Approach},
  journal   = {European Journal of Operational Research},
  volume    = {218},
  number    = {1},
  pages     = {211--229},
  year      = {2012},
  publisher = {Elsevier},
  doi       = {10.1016/j.ejor.2011.09.031}
}

@article{buckinx2005customer,
  author    = {Buckinx, Wouter and Van den Poel, Dirk},
  title     = {Customer Base Analysis: Partial Defection of Behaviorally Loyal Clients in a Non-Contractual {FMCG} Retail Setting},
  journal   = {European Journal of Operational Research},
  volume    = {164},
  number    = {1},
  pages     = {252--268},
  year      = {2005},
  publisher = {Elsevier},
  doi       = {10.1016/j.ejor.2003.12.010}
}

@article{hollmann2022tabpfn,
  title={TabPFN: A Transformer That Solves Small Tabular Classification Problems in a Second},
  author={Hollmann, Noah and M{\"u}ller, Samuel and Eggensperger, Katharina and Hutter, Frank},
  journal={arXiv preprint arXiv:2207.01848},
  year={2022},
  doi={10.48550/arXiv.2207.01848}
}

@article{usmanhamza2024,
  author  = {Usman-Hamza, Fatima E. and Balogun, Abdullateef O. and Amosa, Ramoni T. and Capretz, Luiz Fernando and Mojeed, Hammed A. and Salihu, Shakirat A. and Akintola, Abimbola G. and Mabayoje, Modinat A.},
  title   = {Sampling-based novel heterogeneous multi-layer stacking ensemble method for telecom customer churn prediction},
  journal = {Scientific African},
  volume  = {24},
  pages   = {e02223},
  year    = {2024},
  doi     = {10.1016/j.sciaf.2024.e02223}
}

@inproceedings{warnakulaarachchi2025,
  author    = {Warnakulaarachchi, Claudia and Kumarapathirage, Sapna},
  title     = {Predictive Banking: A Deep Ensemble Customer Churn Prediction Model for Enhanced Customer Retention},
  booktitle = {World Conference on Information Systems and Technologies},
  pages     = {469--483},
  publisher = {Springer Nature Switzerland},
  address   = {Cham},
  year      = {2025},
  doi       = {10.1007/978-3-031-97119-8_36}
}

@article{ahmad2023,
  author  = {Ahmad, Noman and Awan, Mazhar Javed and Nobanee, Haitham and Zain, Azlan Mohd and Naseem, Ansar and Mahmoud, Amena},
  title   = {Customer personality analysis for churn prediction using hybrid ensemble models and class balancing techniques},
  journal = {IEEE Access},
  volume  = {12},
  pages   = {1865--1879},
  year    = {2024},
  doi     = {10.1109/ACCESS.2023.3334641}
}

@article{prokhorenkova2018,
  title={CatBoost: unbiased boosting with categorical features},
  author={Prokhorenkova, Liudmila and Gusev, Gleb and Vorobev, Aleksandr and Dorogush, Anna Veronika and Gulin, Andrey},
  journal={Advances in Neural Information Processing Systems},
  volume={31},
  year={2018},
  doi={10.48550/arXiv.1706.09516}
}

@inproceedings{ke2017,
  title={LightGBM: A highly efficient gradient boosting decision tree},
  author={Ke, Guolin and Meng, Qi and Finley, Thomas and Wang, Taifeng and Chen, Wei and Ma, Weidong and Ye, Qiwei and Liu, Tie-Yan},
  booktitle={Advances in Neural Information Processing Systems},
  volume={30},
  pages={3146--3154},
  year={2017},
  url={https://proceedings.neurips.cc/paper/2017/hash/6449f44a102fde848669bdd9eb6b76fa-Abstract.html}
}

@article{jin2024flexible,
  author    = {Jin, Junwei and Geng, Biao and Li, Yanting and Liang, Jing 
               and Xiao, Yang and Chen, C. L. Philip},
  title     = {Flexible Label-Induced Manifold Broad Learning System for 
               Multiclass Recognition},
  journal   = {IEEE Transactions on Neural Networks and Learning Systems},
  volume    = {35},
  number    = {11},
  pages     = {16076--16090},
  year      = {2024},
  publisher = {IEEE},
  doi       = {10.1109/TNNLS.2023.3291793}
}

@article{jin2022regularized,
  author    = {Jin, Junwei and Qin, Zhenhao and Yu, Dengxiu and Li, Yanting 
               and Liang, Jing and Chen, C. L. Philip},
  title     = {Regularized Discriminative Broad Learning System for Image 
               Classification},
  journal   = {Knowledge-Based Systems},
  volume    = {251},
  pages     = {109306},
  year      = {2022},
  publisher = {Elsevier},
  doi       = {10.1016/j.knosys.2022.109306}
}
\begin{IEEEbiography}
[{\includegraphics[width=1in,height=1.25in,clip,keepaspectratio]{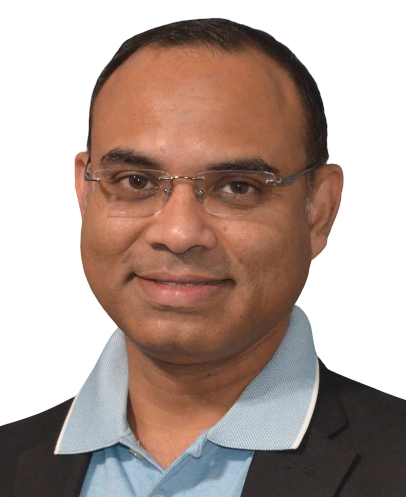}}]
{Joyjit Roy} is a senior technology and program management leader with over 21 years of experience in enterprise digital transformation, cloud modernization, and applied artificial intelligence initiatives. He currently serves in a principal-level leadership role, leading large-scale modernization programs that integrate machine learning, intelligent automation, and Agile delivery frameworks across global enterprises.

His technical and research interests include applied AI and machine learning, natural language processing, multimodal systems, computer vision, agentic AI, edge AI, cybersecurity automation, and intelligent workflow orchestration. His work focuses on translating emerging AI capabilities into scalable enterprise solutions that improve decision-making, operational efficiency, and organizational resilience.

Mr. Roy is a Senior Member of IEEE, a Fellow of the British Computer Society (FBCS), and a Fellow of the Association for Project Management (FAPM). He is an active contributor to the professional and academic community through peer review, technical judging, research publications, and speaking engagements. His professional interests include enterprise AI adoption, governance-driven automation, and technology leadership.
\end{IEEEbiography}

\begin{IEEEbiography}
[{\includegraphics[width=1in,height=1.25in, clip,keepaspectratio]{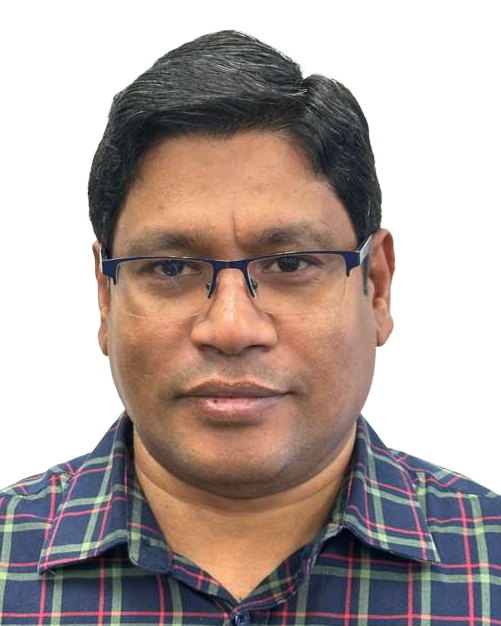}}]{Samaresh Kumar Singh} is a Principal Engineer with over 21 years of experience architecting large-scale distributed systems across edge computing, IoT/IIoT, agentic AI, cloud platforms, and cybersecurity. He specializes in resilient, low-latency architectures for deploying and operating AI/ML workloads across heterogeneous hardware in cloud-edge environments. He has led major platform modernization initiatives, developed distributed compute systems, and driven substantial gains in scalability, reliability, and performance across production infrastructure.

Mr. Singh holds a master’s degree in Computer Software Engineering from Colorado Technical University, USA, and a bachelor’s degree in Computer Engineering from the Institute of Engineering and Management, India. He has contributed to widely adopted open-source initiatives spanning AI/ML frameworks, large language model inference and serving platforms, computer vision and scientific computing libraries, distributed systems, observability infrastructure, performance-critical systems software, and actively engages with technical communities through code contributions, design reviews, and mentorship.

His research interests include distributed edge intelligence, trustworthy AI, IDS/IPS systems, hardware-aware model deployment, and performance and energy-aware orchestration at scale. He is a Senior Member of IEEE, an active technical reviewer, and a mentor in the global engineering community.
\end{IEEEbiography}

\begin{IEEEbiography}
[{\includegraphics[width=1in,height=1.25in,clip,keepaspectratio]{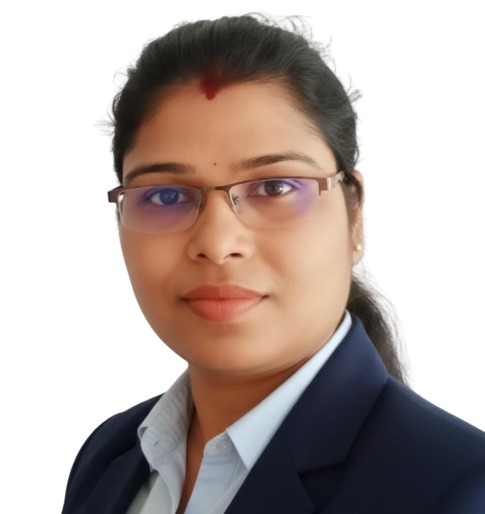}}] {Laxmi Shaw} is a Lead Data Scientist at a multinational financial organization in the United States. From 2023 to 2025, she was a Postdoctoral Scholar in the Department of Information Systems and Analytics at Texas State University and a research collaborator at The University of Texas at Austin, Dell Medical School, where her work focused on adversarial machine learning, large language models, healthcare fraud analytics, Gen AI, computer vision, agentic AI and predictive biomarker modeling using high-performance computing. She has over fifteen years of combined research and industry experience, including roles at Samsung Research and Carrier Corporation, with expertise spanning AI-driven analytics, IoT systems, and digital-twin modeling.

Dr. Shaw received the Ph.D. degree in electrical engineering with a specialization in artificial intelligence and machine learning from IIT Kharagpur, India. She also holds the M.Tech. degree from Jadavpur University and the B.E. degree from Sambalpur University. She is the author of six books and has published over 42 peer-reviewed papers in journals, conferences, and book chapters.

Her research interests include AI/ML security, EEG signal processing, adversarial robustness in LLMs, Siamese networks, and GPU-accelerated healthcare analytics. She is a Senior Member of IEEE, an active reviewer for several IEEE, Springer Nature and Elsevier venues, and a recipient of awards including IEEE Best Paper Awards and the ACM-W Women in Smart Computing Award. Since January 2025, she has been serving as Adjunct Faculty and Research Scientist at Texas A\&M University–Victoria.
\end{IEEEbiography}

\EOD
\end{document}